\documentclass{article}
\usepackage{ijcai25}

\usepackage{times}
\usepackage{soul}
\usepackage{url}
\usepackage[hidelinks]{hyperref}
\usepackage[utf8]{inputenc}
\usepackage[small]{caption}
\usepackage{graphicx}
\usepackage{amsmath}
\usepackage{booktabs}
\usepackage[switch]{lineno}
\usepackage{amsthm}
\usepackage{xcolor}
\usepackage{natbib} 
\usepackage{booktabs}
\usepackage{multirow}
\usepackage{caption}
\usepackage{subcaption}
\usepackage{newfloat}
\usepackage{listings}
\usepackage{float}
\usepackage{amssymb}
\usepackage{graphicx}
\usepackage[ruled,vlined]{algorithm2e}

\title{IGraSS: Learning to Identify Infrastructure Networks from Satellite Imagery by Iterative Graph-constrained Semantic Segmentation}

\author{
    Oishee Bintey Hoque$^{2}$ \and
    Abhijin Adiga$^1$ \and
    Aniruddha Adiga$^1$ \and
    Siddharth Chaudhary$^4$ \and
    Madhav V. Marathe$^{1,2}$ \and
    S. S. Ravi$^{1}$ \and
    Kirti Rajagopalan$^{3}$ \and
    Amanda Wilson$^{1}$ \And
    Samarth Swarup$^{1}$ \\
\affiliations
    $^1$Biocomplexity Institute, University of Virginia, \\
    $^2$Department of Computer Science, University of Virginia\\
    $^3$Department Biomedical Systems Engineering, Washington State University \\
    $^4$Earth System Science Center, University of Alabama in Huntsville\\
}

\begin{document}

\maketitle

\begin{abstract}
Accurate canal network mapping is essential for water management, including irrigation planning and infrastructure maintenance. State-of-the-art semantic segmentation models for infrastructure mapping, such as roads, rely on large, well-annotated remote sensing datasets. However, incomplete or inadequate ground truth can hinder these learning approaches. Many infrastructure networks 
have graph-level properties such as reachability to a source (like canals) or connectivity (roads) that 
can be leveraged to improve these existing ground truth. This paper develops a novel
iterative framework \emph{IGraSS}, combining a semantic segmentation module—incorporating RGB and additional modalities (NDWI, DEM)—with a graph-based ground-truth refinement module. The segmentation module processes satellite imagery patches, while the refinement module operates on the entire data viewing the infrastructure network as a 
graph. Experiments show that IGraSS reduces unreachable canal segments from ~18\% to ~3\%, and training with refined ground truth significantly improves canal identification. IGraSS serves as a robust framework for both refining noisy ground truth and mapping canal networks from remote sensing imagery. We also demonstrate the effectiveness and generalizability of IGraSS using road networks as an example, applying a different graph-theoretic constraint to complete road networks.

\end{abstract}

\section{Introduction}

\begin{figure}[ht]
\centering
  \centering
  \includegraphics[width =\linewidth]{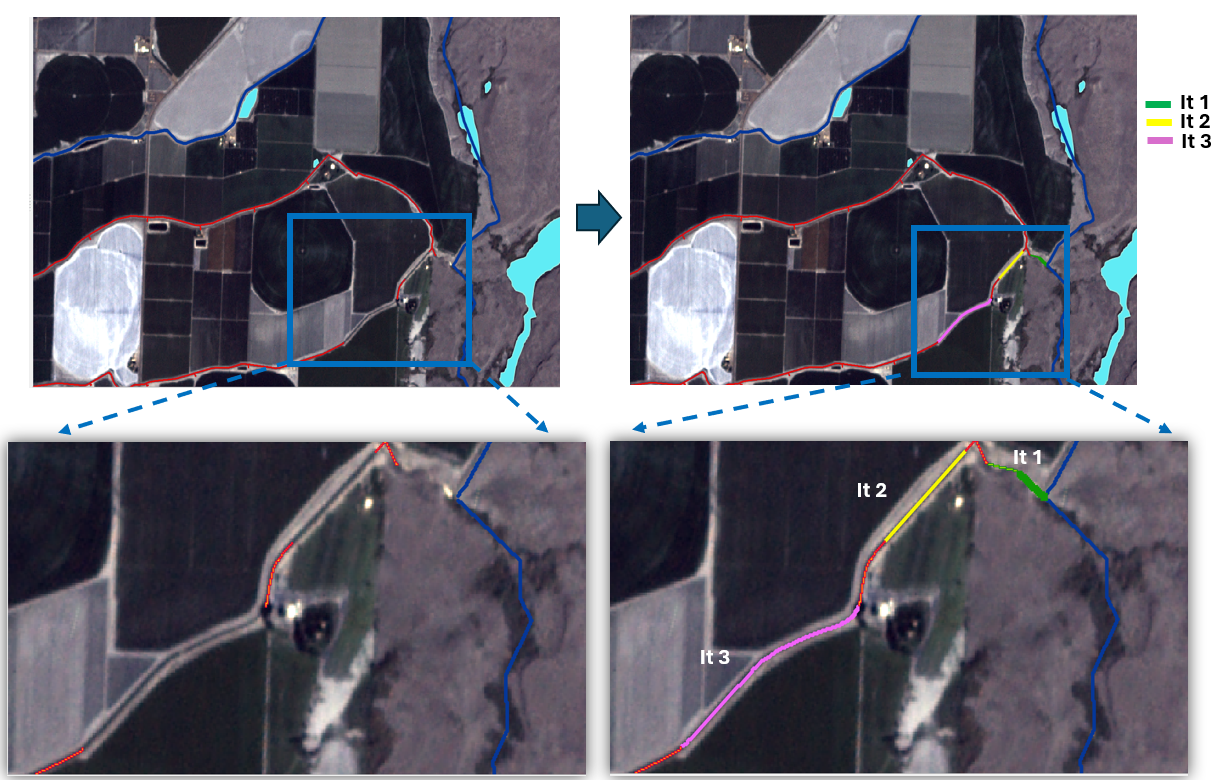}
  \caption{Visualization of Canal Network Completion via IGraSS:
        Blue lines represent reachable canal pixels, while red lines indicate unreachable canal pixels. The images demonstrate gaps in the red canal segments that are iteratively filled. Initially, the green segments connect one of the unreachable red segments with the blue reachable segment, making the upper red segment reachable in Iteration 1 (It~\#1). In the next iteration (It~\#2), yellow segments connect the smaller unreachable red segments. Finally (It~\#3), pink segments connect the remaining bottom segment by filling the gaps, thus making these canals reachable.}
  \label{fig:example}
\end{figure}
Given the growing challenges of water conservation, modern irrigation patterns are shifting towards more cost-effective and water-efficient systems \citep{Blanco2020,FAN2023107565}. Furthermore, increasing pressures from drought, rising operational costs of canal infrastructure, and the decreasing cost of canal technology are driving many irrigation districts toward modernization~\citep{Belt2009,creaco2023,FAN2023107565,creaco2023}. Modernizing irrigation infrastructure requires accurate knowledge of existing canal networks, but manual mapping is slow and outdated maps lack completeness \citep{HOSSEINZADE2017177}.{We tackle this challenge by using high-resolution remote sensing to automatically extract canal networks, which will inform efforts on water management and infrastructure planning. This contributes to sustainable development and aligns with United Nations Sustainable Development Goals~12 (SDG 12) by promoting responsible resource use in agriculture and related sectors~\citep{UN2015}}.


Irrigation canal mapping by agencies typically relies on a labor-intensive GIS process, where experts manually annotate lines and polygons \citep{Belt2009,Archuleta_Terziotti_2023}. This approach often results in noisy, incomplete labels (See Fig \ref{fig:example}) and requires updates over time as infrastructure evolves—for example, open canals being converted to closed pipe systems or covered with solar panels for efficiency \citep{Loureiro2024}. An automated remote-sensing approach can streamline updates and help assess water and energy efficiency improvements. While road network extraction is well-studied~\citep{abdollahi2020deep} with high-quality benchmarks~\citep{spacenet,deepglobe2018}, datasets for other infrastructure networks, such as irrigation canals, remain limited, often containing insufficient data and noisy, incomplete annotations.



Our approach leverages graph-theoretic properties (e.g., reachability, connectivity, planarity) to address noisy or incomplete infrastructure annotations. In irrigation networks, for instance, every canal segment should be reachable from a water source. Similar constraints apply to road and power networks. Unlike prior work using topological representations for road extraction, our method integrates graph constraints by aggregating segmentation outputs across multiple images. To effectively apply graph constraints such as reachability and connectivity, segmentation outputs must be aggregated across multiple images. Incorporating the Normalized Difference Water Index (NDWI)~\citep{xu2006modification} and a digital elevation model (DEM) along with RGB significantly improves segmentation performance \citep{nagaraj2024extraction} by enhancing feature representation. Beyond segmentation, we propose a method to refine ground truth, which is often fragmented or disconnected. Using graph-theoretic properties, we correct these inconsistencies and show that improved annotations further boost model performance (See Figure~\ref{fig:example}).

\paragraph{Contributions.} Our contributions in this work are: (i) a framework
that allows the combination of learning with optimization/constraint satisfaction methods
through the use of \emph{pseudo-labels}, (ii) a refined set of metrics, called 
\emph{r-neighborhood} metrics, which are more suitable for evaluating the performance
of semantic segmentation problems like the ones studied here, (iii) a demonstration of
improved performance in canal network identification (using reachability constraints), (iv) demonstration of generalizability using road network completion as an example, optimizing pairwise distances under a graph-theoretic constraint. (v) {We provide refined canal data for the state of Washington that is more connected and less noisy, along with the accompanying code for research purposes: https://github.com/oishee-hoque/IGraSS.}

{\paragraph{Team.}
This work is the result of an interdisciplinary collaboration between computer scientists, an agro-ecosystems modeler with expertise in water and agricultural resource management, and an earth science and remote sensing expert.}

\section{Related Work}
 
\citet{liu2022survey} provide a comprehensive review of infrastructure network
extraction using deep learning, particularly in the context of roads.) We use
popular models from this literature as backbone networks for our semantic
segmentation module. Similar
approaches have also been applied to related problems such as crack
detection, blood vessel segmentation, abnormality in anatomical structures,
and extracting power systems~\citep{ganaye2018semi,cheng2021joint,ren2022automated}.

\smallskip
\noindent
\textbf{Graph-based segmentation methods} have been explored for road network inference and related tasks. Our work is most closely related to RoadTracer~\citep{bastani2018roadtracer}, which iteratively constructs road graphs using dynamic labels. However, unlike our approach, which refines labels based on global constraints (e.g., reachability), their method relies on a CNN-based decision function constrained to local patches.

Other relevant works include Sat2Graph~\citep{he2020sat2graph}, which encodes road graphs as tensors for deep aggregation networks, and GA-Net~\citep{chen2022ga}, which integrates segmentation with geometric road structures to enhance connectivity. Additionally, road boundaries detected via traditional filtering serve as inputs to a deep learner with D-LinkNet architecture~\citep{Zhou2018DLinkNetLW}.
\citet{cira2022improving} use an inpainting approach as a postprocessing
technique to link unconnected road segments. \citet{dualGraph}  utilize hypergraphs to capture high-order and long-range 
relationships among roads, incorporating various pretext tasks for optimization and demonstrating significant 
improvements across multiple datasets, tasks, and settings. None of these
approaches use global graph constraints/optimization.


\smallskip
\noindent
\textbf{Mapping water bodies} from remote-sensed data is an active area in
remote sensing. Various unsupervised and supervised methods have been
used. A prominent approach is the use of water
indices such as normalized difference water index~(NDWI). The use of deep
learning methods in this context is an emerging area of research, with several papers using
standard segmentation techniques. (See \citet{nagaraj2024extraction} for an
extensive review.)  \citet{gharbia2023deep} highlights challenges such as
lack of quality data and variations in water body types.
\citet{li2022accurate} consider the extraction of natural water bodies from
binarized NDWI images. They apply a connected-component method followed by
an analysis of shape and spectral characteristics to assign a confidence
value to each water body. This is subsequently used to train peer networks.
\citet{yu2023boundary} address the problem of fine-grained extraction of
water bodies, where the challenge is to accurately detect the boundaries of
water bodies. They propose a novel boundary-guided semantic context network
in this regard. In our work, the emphasis is more on the accuracy of the
network structure of the canal network.

\newcommand{\grid}{G}
\newcommand{\gtg}{G_{gt}}
\newcommand{\cg}{G_{c}}
\newcommand{\pcg}{H_{c}}
\newcommand{\lab}{\ell}
\newcommand{\labr}{\gamma}
\newcommand{\reach}{\textsc{NetReach}}
\newcommand{\fs}{f_s}
\newcommand{\thresh}{\tau}
\newcommand{\sti}{\mathcal{S}}
\newcommand{\stic}{\mathbf{S}}
\newcommand{\rtp}{\mbox{$r\text{TP}$}}
\newcommand{\rfp}{\mbox{$r\text{FP}$}}
\newcommand{\rfn}{\mbox{$r\text{FN}$}}
\newcommand{\rp}{\mbox{$r\text{P}$}}
\newcommand{\rr}{\mbox{$r\text{R}$}}
\newcommand{\rf}{\mbox{$r\text{F1}$}}
\newcommand{\rd}{\mbox{$r\text{I}$}}
\newcommand{\neigh}{\mathcal{N}_r}

\begin{figure*}[ht]
\centering
  \centering
  \includegraphics[width=\textwidth]{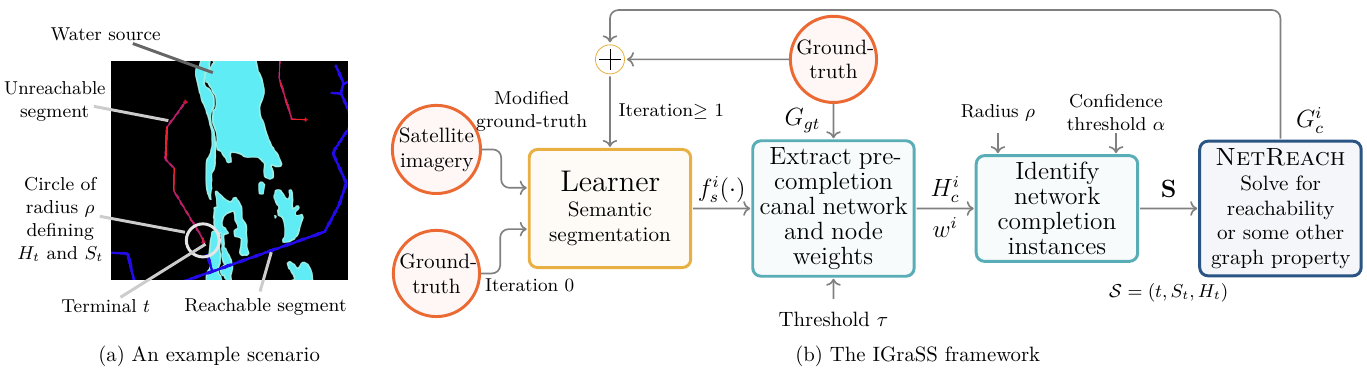}
  \caption{An example patch with gaps in data is shown in~(a). In~(b), an
  outline of our framework is shown.}
  \label{fig:framework}
\end{figure*}
\section{Proposed Framework}
\subsection{Preliminaries}
We use the problem of identifying irrigation canals from satellite images
as the main running example in this paper.
However, the framework can also
be applied to other problems, such as identifying road networks.

We assume that the study region is overlaid with a grid where each
grid cell corresponds to a pixel of a satellite image. Let~$\grid$ denote
the graph induced by this grid where~$V(\grid)$ denotes the set of grid
cells. Two nodes~$u,v\in V(\grid)$ are adjacent if and only if they share
a side or a corner. This corresponds to the Moore neighborhood. The set of
edges is denoted by~$E(\grid)$. The \emph{ground-truth canal
network}~$\gtg$ is a subgraph of~$\grid$ induced by~$V(\gtg)$, the set of
grid cells identified as canal pixels. Further, each~$v\in V(\grid)$ is
assigned a label~$\lab(v)\in\{0,1\}$, where~$0$ denotes a non-canal
node and~$1$ denotes a canal node. For all canal
nodes~($\lab(v)=1$), the label~$\labr(v)=1$ implies that it is reachable
from a water source, while~$0$ means otherwise. Henceforth, for brevity, we
use the phrase ``$v$ is reachable'' to denote that the node is reachable
from a water source.

\subsection{Problems}
\paragraph{Inference goal.} Under the assumption that the ground-truth
canal network is incomplete (but not erroneous), the goal is to infer the
canal network~$\cg$, an induced subgraph of~$\grid$ such that all
nodes in~$V(\cg)\supseteq V(\gtg)$ are reachable; i.e.,~$\forall v\in
V(\cg),\, \labr(v)=1$.

and the trained model, the objective is to identify all the edges of the canal network~$\cg$.

\subsection{Approach}
Training data for neural network-based semantic segmentation approaches
divide large satellite images into \emph{patches}, where a patch is 
a small rectangular part of the large image. Typical patch sizes are 256$\times$256
pixels or 512$\times$512 pixels, whereas the image as a whole can be, say,
$\approx$~20,000$\times$30,000 pixels in size. (See the Experimental Evaluation
section for details regarding the data). While the classifier can be trained using each
patch as a training input and the intersection of~$V(\gtg)$ with that patch
as the corresponding expected output for training, it is important to note
that the overall graph constraints (reachability, connectedness) are \emph{global}
constraints. These cannot be directly incorporated into the learning process
by augmenting the objective function, as is customary in constrained
learning problems.

Figure~\ref{fig:framework}(a) shows a toy example where red-marked canals are
not reachable from any water sources. The dark blue canals are reachable from
the water source. Our goal is to identify these disconnected canal segments
and connect them to nearby water sources or to reachable (blue) canals. 
We therefore develop a new, iterative approach composed of (i)~a learner that infers a
canal network given ground-truth data and satellite imagery after training for a number of intermediate epochs, and (ii)~a
\emph{constraint solver} algorithm that modifies the training data after each iteration
by adding positive \emph{pseudo-labels} to chosen pixels. A positive pseudo-label refers
to changing the label of a pixel from~$\lab(v)=0$ to ~$\lab(v)=1$ to satisfy the constraint.
In the canal network identification problem,
the constraint solver is a network completion algorithm to satisfy the reachability
constraint. See the outline in Figure~\ref{fig:framework}. In each iteration, step (ii) 
modifies the expected outputs in the training data for the next round of training.
The canal network is initially set to the ground-truth network, i.e.,~$\cg^0=\gtg$, and
is then modified in step (ii) of each iteration.
The following are the steps in each iteration~$i \geq 1$. 

\noindent
\textbf{Learner.} The learner's objective is to provide the likelihood
that a node in the grid graph belongs to a canal. It is trained using
modified ground-truth obtained from~$\cg^{i-1}$ for~$\lambda_i$ epochs. For
the purpose of training in this iteration, for any
node~$v$,~$\lab(v,i-1)=1$ (i.e., $v$ is a canal pixel
in iteration $i-1$) if and only if $v\in
V(\cg^{i-1})$. Let~$\fs^i(v)$ denote the output of the learner. Note that
the output depends on the learning methodology.

\noindent
\textbf{Pre-completion network.} Given the learner's output, ground-truth
network~$\gtg$, and a user-specified threshold~$\tau$, a canal network
graph~$\pcg^i$ is computed. Firstly, a likelihood~$w^i(v)\in[0,1]$ is
computed from~$\fs^i(v)$. 
We construct a \emph{pre-completion
canal network}~$\pcg^i$ as a graph induced by the nodes that satisfy the following condition: (i)~$v\in
V(\gtg)$ (ground-truth) or (ii)~$w^i(v)\ge\tau$.

\noindent
\textbf{Network-completion instances.}
Given the pre-completion network~$\pcg^i$, its node set is partitioned into
reachable and unreachable nodes. A set of candidate instances to apply
network completion, called \emph{network completion instances}, are
identified. Each such instance~$\sti$ consists of a tuple~$(t,S_t,H_t)$
where~$t$ is an unreachable \emph{terminal},~$S_t=\{s_1,s_2,\ldots\}$
is a collection of reachable nodes called \emph{sources}, and $H_t$, 
referred to as the \emph{$t$-local graph}, is a subgraph of the grid
graph~$G$ containing~$\{t\}\cup S_t$, where each node~$v\in V(H_t)$ has 
weight~$1/w^i(v)$ if~$1/w^i(v)>\alpha$~(a confidence 
threshold),~$0$ otherwise (See Algorithm~\ref{alg:edge}). Only the nodes of~$V$ will be used for 
extending~$t$ to a reachable node. More implementation details are given in the Appendix. Let~$\stic=\{\sti_1,\sti_2,
\ldots\}$ be the collection of such instances.

\noindent
\textbf{Identifying a network-completion instance.} Given~$\pcg^i$,
using morphological thinning \cite{Fisher2003}
end points of canal segments are 
identified. 
An end point~$t$ is a terminal if it is not reachable. All
pixels at distance~$\rho$~(user-specified) from~$t$ form the node set 
of its local graph~$H_v$. The source set of~$v$ is the set of all end 
points in the local graph that are reachable.

\noindent
\textbf{Network reachability computation.}
For each instance~$\sti=(t,S_t,H_t)\in \stic$, the objective is to
find a minimum weighted shortest path from~$t$ to~$S_t$ in~$H_t$. 
This involves converting the node-weighted~$H_t$ to an edge-weighted
graph (See Algorithm \ref{alg:subgraph}), followed by the application of Dijkstra's~\citeyearpar{dijkstra1959note} shortest path
algorithm. More implementation details is given in the Appendix \ref{sec:ap_framework} included in the Online Supplementary Material.

\subsection{Framework Modules}
\label{sec:framework}



\paragraph{Reachable and Unreachable Nodes.} Given water source indices as \( F \), Algorithm~\ref{alg:directly_connected} identifies canal pixels directly connected to these sources. Given a binary matrix \( M \) representing the canal network (pre-completion \( \pcg^i \)), the algorithm creates a boolean mask \( B \), marking the locations of \( F \). It then applies a convolution with an 8-connectivity kernel to determine the set of directly connected canal pixels, \( C \), which are classified as reachable. A breadth-first search (BFS) expands from \( C \) to find all connected canal pixels, forming the set \( R \) of reachable pixels. The remaining canal pixels in \( M \) that are not in \( R \) are considered non-reachable \( U \).
\begin{algorithm}[h]
\footnotesize
\caption{Directly Connected Canal Nodes}
\KwIn{$M$: A binary matrix of size $m \times n$ \\
      $F$: A set of water index pairs $(i, j)$}
\KwOut{$C$: Set of directly connected 1s}

Create boolean mask $B$ of size $m \times n$ with $B[i,j]=1$ if and only if $(i,j) \in F$.

Define kernel $K$:
\[
K = \begin{bmatrix}
1 & 1 & 1 \\
1 & 0 & 1 \\
1 & 1 & 1
\end{bmatrix}
\]

$V \gets $ Convolve $B$ with $K$\;

Initialize empty set $C$\;
\ForEach{$(i,j) \in M$}{
    \If{$V[i,j] > 0$ \textbf{and} $M[i,j] = 1$}{
        Add $(i,j)$ to $C$\;
    }
}

\Return{$C$}\;
\label{alg:directly_connected}
\end{algorithm}







\paragraph{Terminals.} Given the set of unreachable nodes ~$U$, we identify the terminal nodes. For each point $p \in U$, the algorithm examines its 8-connected neighborhood defined by the directions $\Delta = \{(0, \pm1), (\pm1, 0), (\pm1, \pm1)\}$. A point is classified as a terminal if it has one or fewer neighbors within the set $U$. The algorithm maintains a set $V$ of visited points to avoid redundant computations and it returns the set of terminals $E$.

\begin{algorithm}[h]
\footnotesize
\caption{Edge Point Processing}
\KwIn{$E$: Set of terminal points \\
      $w^i(v)$: Likelihood matrix \\
      $\pcg^i$: Pre-completion network \\
      $\rho$: Sampling radius \\
      $\alpha$: Confidence threshold}
\KwOut{$X_r$: Resultant matrix}

$X_r \gets \pcg^i$ \tcp{Initialize with pre-completion network} 
$(n, m) \gets$ size of $\pcg^i$ \tcp{Extract matrix dimensions} 

\ForEach{$p \in E$}{
    $N_p \gets$ \textbf{GetNeighbors}$(p, \rho, (n, m))$\;
    \ForEach{$n \in N_p$}{
        \If{$w^i[n] > \alpha$ \textbf{and} $X_r[n] = 0$}{
            $X_r[n] \gets \lfloor 1 / w^i[n] \rfloor$\;
        }
    }
}

\Return $X_r$\;

\BlankLine
\SetKwFunction{FGetNeighbors}{GetNeighbors}
\SetKwProg{Fn}{Function}{:}{}
\Fn{\FGetNeighbors{$p, \rho, (n, m)$}}{
    $(x, y) \gets p$\;
    $N \gets \emptyset$\;
    \For{$dx \in [-\rho, \rho]$}{
        \For{$dy \in [-\rho, \rho]$}{
            \If{$dx \neq 0$ \textbf{or} $dy \neq 0$}{
                $(nx, ny) \gets (x + dx, y + dy)$\;
                \If{$0 \leq nx < n$ \textbf{and} $0 \leq ny < m$}{
                    $N \gets N \cup \{(nx, ny)\}$\;
                }
            }
        }
    }
    \Return $N$\;
}
\label{alg:edge}
\end{algorithm}
\paragraph{Source-Terminal Pairs.}  
For each terminal \( t \in E \), we identify source points \( S_t \) within a radius \( \rho \). The sources include water source edge points (those with fewer than eight neighbors) and reachable canal nodes. Using an approach similar to Algorithm~\ref{alg:terminals}, we first detect water source edges and then determine source-terminal pairs by selecting all source points \( p \in S \) that satisfy \( \|p - t\|_2 \leq \rho \).
\begin{algorithm}[!h]
\footnotesize
\caption{Directed Subgraph Around Terminal}
\KwIn{$M_{m\times n}$: Matrix representing the canal network \\
      $t = (t_x, t_y)$: Terminal coordinates \\
      $r$: Radius}
\KwOut{$G = (V,E)$: Directed graph}

$R \gets \{(i,j) \mid |i-t_x| \leq r, |j-t_y| \leq r, 0 \leq i < m, 0 \leq j < n\}$\;
$\Delta \gets \{(0, \pm1), (\pm1, 0), (\pm1, \pm1)\}$\;
$V, E \gets \emptyset$\;

\ForEach{$(i,j) \in R$ \textbf{where} $M[i,j] > 0$}{
    $V \gets V \cup \{(i,j,1), (i,j,2)\}$\;
    $E \gets E \cup \{((i,j,1), (i,j,2), M[i,j])\}$\;
    
    \ForEach{$(d_x, d_y) \in \Delta$}{
        \If{$(i+d_x, j+d_y) \in R$ \textbf{and} $M[i+d_x, j+d_y] > 0$}{
            $E \gets E \cup \{((i,j,2), (i+d_x, j+d_y,1), 0),$\\
            \hspace{3em} $((i+d_x, j+d_y,2), (i,j,1), 0)\}$\;
        }
    }
}

\Return $G = (V, E)$\;
\label{alg:subgraph}
\end{algorithm}

\subsection{Metrics}

Alongside conventional metrics—Precision~(P), Recall~(R), F1 Score~(F1), and Intersection over Union~(IoU)—\emph{we introduce parameterized metrics} to address width inconsistencies in thin-structure segmentation. These metrics account for minor spatial misalignments in single-pixel annotations, such as canal networks represented by shapefile line segments, which may not perfectly align with the ground truth. Conventional metrics may underestimate performance in such cases, even when the model captures the overall structure. To address this, we define an $r$-neighborhood ($\neigh(i,j)$) around each pixel $(i,j)$, allowing for small spatial deviations. Let $p_{i,j} \in \{0,1\}$ be the predicted value at $(i,j)$ and $y_{k,l} \in \{0,1\}$ the ground truth at $(k,l)$.

\noindent
\textbf{$r$-Neighborhood True Positives (\rtp)}
is the number of predicted positive pixels that lie
within the $r$-neighborhood of the actual positive pixels:
$
\text{\rtp} = \sum_{i=1}^{N} \sum_{j=1}^{M} \max_{k,l \in \neigh(i,j)}
\left( y_{k,l} \right) \cdot p_{i,j}\,.
$

\noindent
\textbf{$r$-Neighborhood False Positives (\rfp)}
are the number of predicted positive pixels that do not lie within the
$r$-neighborhood of any actual positive pixels:
$
\text{\rfp} = \sum_{i=1}^{N} \sum_{j=1}^{M} p_{i,j} \cdot \big( 1 -
\max_{k,l \in \neigh(i,j)} y_{k,l} \big)\,.
$

\noindent
\textbf{$r$-Neighborhood False Negatives (\rfn)}
are the number of actual positive pixels for which there are no predicted
positive pixels within the $r$-neighborhood:
$
\text{\rfn} = \sum_{i=1}^{N} \sum_{j=1}^{M} y_{i,j} \cdot \big( 1 -
\max_{k,l \in \neigh(i,j)} p_{k,l} \big)\,.
$

\begin{figure}[!ht]
    \centering
    \includegraphics[width=.9\linewidth]{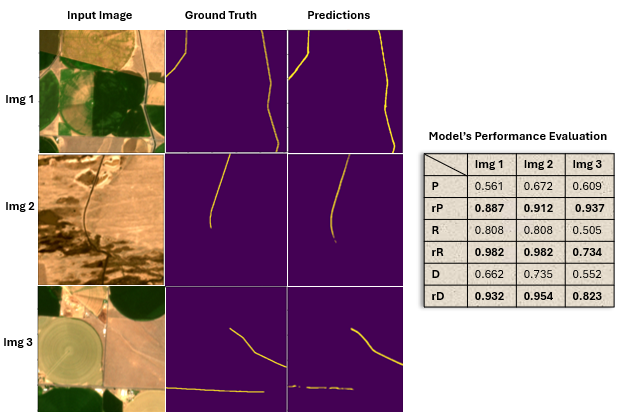}
\caption{Comparison of conventional and refined metrics in evaluating thin-structure segmentation. These examples illustrate how our refined metrics more accurately assess model performance by reducing reliance on pixel-level precision. While the model effectively captures overall structures, conventional metrics fail to fully reflect this capability. In contrast, our refined metrics provide a more nuanced evaluation, better representing the model’s ability to identify key structural elements.}
    \label{fig:refined_metrics}
\end{figure}

The $r$-IoU ~(\rd), $r$-precision~(\rp), $r$-recall~(\rr), and $r$-F1
score~(\rf) are
similar to their conventional counterparts with TP, FP, and FN replaced by
$\rtp$, $\rfp$, and $\rfn$ respectively. The usefulness of these metrics is
demonstrated in Figure~\ref{fig:refined_metrics}.

\section{Experimental Setup}
\paragraph{Canal Network Dataset.}
We used PlanetScope (2020–2023) 3m-resolution RGB imagery to map irrigation canals in central Washington~\citep{NASA_CSDAP}. NDWI was computed from Green and NIR channels, while USGS 3DEP DEM (1m)~\citep{USGS_3DEP} was resampled to 3m and used alongside NDWI and RGB. Canal waterway data came from the National Hydrography Dataset (NHD)~\citeyearpar{usgs_nhd}. To prepare the dataset, we merged imagery tiles into a $\approx$ 20,000$\times$30,000 tile per year and divided them into non-overlapping 512$\times$512 patches. We filtered patches with over 30\% black pixels and excluded mask patches with fewer than 0.5\% canal pixels. We have total of ~30,000 512x512x5 (RGB, NDWI, DEM) patches in our dataset.

For experiments, we created two distinct sets (Set~1 and Set~2) of spatially separated training, validation, and test data. The test set was 20\% of the data, while 80\% was used for training and validation. We employed 5-fold cross-validation and ran IGraSS to track average performance metrics. After tuning hyperparameters (iterations, epochs, $\rho$, etc.), we re-ran IGraSS on training data and evaluated on the test sets.

\paragraph{Road Network Data}
Following RoadTracer~\cite{bastani2018roadtracer}, we obtained 60 cm/pixel satellite imagery of New York City (24 sq km) from Google Maps and merged the tiles. We extracted the road network from OpenStreetMap, converted coordinates to match the imagery, and generated road masks. To create partial road maps, we iteratively removed $\alpha$ road segments of random length $\beta$ from a predefined list (e.g., $\beta \in \{20,30,50,100\}$). Further details, including train-test split, are in Appendix \ref{sec:Dataset}.


\begin{table*}[h]
\footnotesize
\centering
\begin{tabular}{l|r|l|r|r|r|r|r|r|r|r}
\toprule
\textbf{Model} & \textbf{Test Set} & \textbf{w/ or w/o} & \textbf{P} & \textbf{\rp} & \textbf{R} & \textbf{\rr} & \textbf{F1} & \textbf{\rf} & \textbf{I} & \textbf{\rd} \\
\midrule
\multirow{4}{*}{\textbf{ResUnet}}  
& \multirow{2}{*}{1} & w/o & 0.591 & 0.838 & 0.540 & 0.612 & 0.564 & 0.708 & 0.531 & 0.601 \\
& & w & 0.643 & 0.874 & 0.589 & 0.668 & 0.615 & 0.757 & 0.546 & 0.620 \\
\cline{2-11} \addlinespace[.1em]
& \multirow{2}{*}{2} & w/o & 0.549 & 0.802 & 0.521 & 0.599 & 0.535 & 0.686 & 0.560 & 0.630 \\
& & w & 0.648 & 0.878 & 0.587 & 0.664 & 0.617 & 0.756 & 0.585 & 0.660 \\
\midrule
\multirow{4}{*}{\textbf{Deeplabv3+}} & \multirow{2}{*}{1} & w/o & 0.613 & 0.805 & 0.580 & 0.765 & 0.596 & 0.780 & 0.420 & 0.600
                                                         \\
                                                         &                    
                                                         & w & 0.644 & 0.835 & 0.610 & 0.790 & 0.626 & 0.808 & 0.440 & 0.620 \\
                                                         \addlinespace[.1em]\cline{2-11}\addlinespace[.1em]
                                     & \multirow{2}{*}{2} 
                                                          & w/o & 0.605 & 0.798 & 0.570 & 0.758 & 0.587 & 0.773 & 0.415 & 0.590 \\
                                                          
                                                          &                    
 
                                                          & w & 0.636 & 0.829 & 0.600 & 0.785 & 0.617 & 0.800 & 0.435 & 0.610 \\
\midrule
\multirow{4}{*}{\textbf{SwinTransformer}}  
& \multirow{2}{*}{1} & w/o & 0.775 & 0.850 & 0.765 & 0.835 & 0.770 & 0.842 & 0.720 & 0.805 \\
& & w & \textbf{0.820} & \textbf{0.900} & \textbf{0.810} & \textbf{0.885} & \textbf{0.815} & \textbf{0.892} & \textbf{0.760} & \textbf{0.850} \\
\cline{2-11} \addlinespace[.1em]
& \multirow{2}{*}{2} & w/o & 0.770 & 0.845 & 0.760 & 0.830 & 0.765 & 0.838 & 0.715 & 0.800 \\
& & w &\textbf{ 0.815} & \textbf{0.895} & \textbf{0.805} & \textbf{0.880} & \textbf{0.810} & \textbf{0.887} & \textbf{0.755} & \textbf{0.845} \\
\midrule

\bottomrule

\end{tabular}
\caption{Performance comparison of IGraSS against baseline models on Canal Network datasets. Here `w/' indicates the using IGraSS in conjunction with the baseline models, while `w/o' represents training the baseline models without IGraSS for the same number of epochs.}
\label{tab:merged_performance}
\end{table*}
\section{Results}
\subsection{Irrigation canals}
Our evaluation focuses on two main aspects: (1) comparing IGraSS's performance against the state-of-the-art models used as 
learners in our framework, and (2) assessing its ability to complete canal networks given reachability constraints. We also perform extensive experiments under various parameter settings 
to provide a comprehensive evaluation. 

\paragraph{Segmentation Baseline Networks.} For our experiments, we select three popular state-of-the-art models to 
serve as the Learner in our framework: DeepLabV3+~\citep{deeplab}, ResNet50~\citep{he2015deep}, ResUNet~\citep{resunet}, 
and Swin Transformer~\citep{he2022swin}. To the best of our knowledge, no work has been done on irrigation canal identification using deep learning and remote sensing images. Therefore, we use state-of-the-art models' performances as our baselines.

To assess the impact of iterative ground truth refinement on the model's overall performance, we conduct a systematic 
analysis. Intuitively, breaks or inconsistencies in the ground truth should negatively affect the model's performance. 
Conversely, as the quality of the ground truth improves through our iterative process, we expect to see a positive 
impact on the model's performance. This analysis aims to verify this hypothesis and quantify the relationship between 
ground truth refinement and model's performance. 

To establish a fair comparison, we train each baseline model independently for same number of epochs without 
implementing our framework. The training setup for these baselines is identical to the one used within our framework, 
ensuring consistency in our evaluation.
The key distinction lies in the treatment of ground truth data. In our proposed method, the ground truth is updated 
after each iteration using the output from our framework. In contrast, the baseline models are trained using the 
original, unmodified ground truth throughout the entire process. We ran IGraSS framework for 5 iterations with a radius 
of 100 and an initial confidence threshold $\alpha$ of 0.2, which was later reduced to 0.01 for optimal result.  

 

\paragraph{Performance evaluation by refining Ground Truth.}

Table~\ref{tab:merged_performance}
presents the performance evaluation on the two different test sets of the Canal Network Dataset across the 3 models used for training. 
 The reported results in Table \ref{tab:merged_performance}, demonstrate that training the model using refined ground truth from IGraSS significantly enhances the performance of all models across all metrics. Swin Transformer outperformed other models, as it breaks the image into small patches, computing self-attention locally while enabling cross-window connections. This helps capture the overall canal structure relationships within patches. For Swin Transformer, model trained with the IGraSS's refined ground truth improves precision from 0.775 to 0.820 (5.8\% increase), recall from 0.765 to 0.810 (5.2\% increase), F1-score from 0.770 to 0.815 (\textbf{5.8\% increase}), and IoU from 0.720 to 0.760 (5.6\% increase) on Test Set 1. Similar trends are observed in Test Set 2, with improvements of 5.8\% in precision, 5.3\% in recall, 5.9\% in F1-score, and 5.6\% in IoU.
The refined ground truth also improves Deeplabv3+ performance by ~5\% and enhances ResUnet performance ~10\% across all metrics. These improvements are evident not only in our proposed metrics but also in conventional scores showing better ground truth refined by IGraSS, boost model's performance.

\begin{figure}[h]
    \centering
    \includegraphics[width=\linewidth]{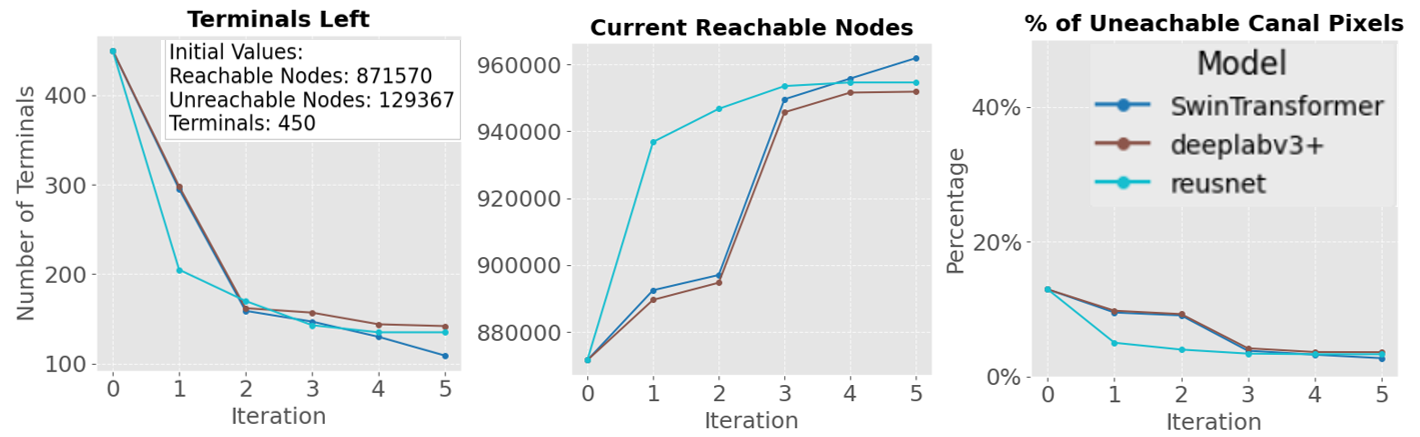}
    \caption{Network Completion Analysis with IGraSS.}
    \label{fig:network}
\end{figure}
\paragraph{Network completion assessment.}
Figure \ref{fig:network} presents a quantitative analysis of the number of canal pixels that get connected in each iteration. All three models achieved similar results in connecting the terminals after 5 iterations. As we refine the main ground truth by connecting only the terminals to the nearest water sources, the number of reachable canals increases with each iteration, while the number of unreachable canal pixels and terminals decreases. In both cases IGraSS was able to reduce down the unreachable canals from ~$(18-15)\%$ to $(5-3)\%$.
Upon manual analysis of the terminals that remained unreachable, we found that in most cases, there were either no water sources within the selected radius or the corresponding images did not contain any visible canals at all. This lack of visible canals might have led the neural network to fail in predicting any canal pixels.




\subsection{Ablation Study}

\begin{table}[h]
\footnotesize
    \centering
    \begin{tabular}{lr|r|r|r}
        \toprule
        \textbf{Model} & \textbf{RGB} & \textbf{+N} & \textbf{+D} & \textbf{+N+D} \\
        \midrule
        DeepLab  & 0.290 & 0.406 & 0.348 & 0.435 \\
        ResUnet  & 0.304 & 0.425 & 0.365 & 0.556 \\
        Swin     & 0.523 & 0.731 & 0.627 & \textbf{0.785} \\
        \bottomrule
    \end{tabular}
    \caption{Performance comparison of different models with various input modalities (IoU only).}
    \label{tab:performance_comparison}
\end{table}

\begin{figure}[ht]
    \centering
    \includegraphics[width=\linewidth]{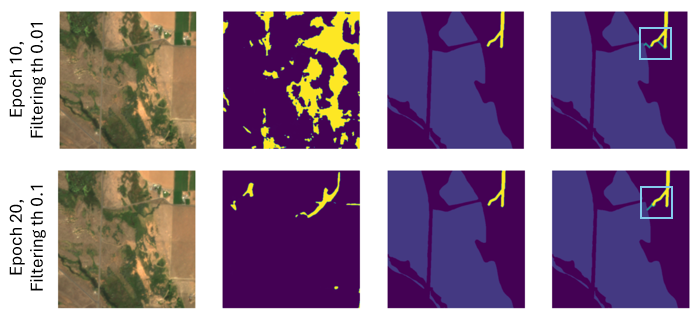}
    \caption{Error Results}
    \label{fig:error_results_main}
\end{figure}

\paragraph{Effect of adding NDWI and DEM.} In table \ref{tab:performance_comparison}, we evaluated the performance of four deep learning models—DeepLabv3+, ResNet50, ResUnet, and SwinTransformer—using different input modalities: RGB, RGB+NDWI (N), RGB+DEM (D), and RGB+NDWI+DEM (N+D). Starting with RGB as the baseline, adding NDWI improved performance by 35-40\%, highlighting the significance of vegetation indices in segmentation. Incorporating DEM resulted in a 10-20\% improvement, indicating that elevation data contributes useful topographic information. Combining both NDWI and DEM led to the highest boost of 45-50\%, demonstrating their complementary benefits. 
\paragraph{Effect of Framework Parameters.}
The IGraSS framework's performance is influenced by parameters such as intermediate epochs, radius $\rho$, threshold $\tau$, and confidence threshold $\alpha$. We experimented with $\rho \in \{20, 50, 100, 150\}$, $\alpha \in \{0.3, 0.2, 0.1, 0.01\}$, and epochs $\in \{10, 20, 30\}$ using SwinTransformer, DeepLabv3+, and ResUnet, evaluated over 5 iterations with K-Fold cross-validation. Lower $\alpha$ (0.01) with fewer epochs (10) reduces unreachable canals but introduces noise, while moderate $\alpha$ (0.1) with more epochs (20) yields cleaner results. Thresholds between 0.2 and 0.1 were generally effective. The number of epochs is crucial—too few with low $\alpha$ cause noise, while extended training risks learning noisy ground truth. An adaptive approach, lowering $\alpha$ after sufficient training, may be beneficial. Due to space limit, extensive parameter sensitivity analysis is presented in Appendix included in Online Supplementary Material (Tables~\ref{tab:parameter} and \ref{tab:parameter_tune}; further details are in Section~\ref{sec:experiments}).
\paragraph{Error Analysis}
As discussed in our parameter analysis, selecting appropriate parameters is crucial to avoid erroneous connections. IGraSS's focus on connecting points via the shortest path helps minimize errors when adding new data to the ground truth. Directly using the neural network output would have introduced significant noise to the ground truth, which our adaptive thresholding process helps mitigate. However, as illustrated in Figure \ref{fig:error_results_main}, unwanted connections may still occur if the right parameters are not chosen.
\noindent

\subsection{Road Networks}

To further demonstrate our framework's effectiveness and generalizability, we use it to complete road 
networks under a different graph-theoretic constraint; the objective in this case is to minimize the distance (on the network) between any pair of a user-specified set of points on the road network. 
We would like to emphasize here that our objective is to demonstrate the effectiveness of our IGraSS
framework in satisfying graph-based constraints and not to show improved performance on these 
road networks which are already of high quality.

\noindent\textbf{Problem Statement:} Given a complete road network \( G = (V, E) \), we introduce random gaps by removing a subset of edges \( E_r \subset E \), resulting in an incomplete network \( G' = (V, E \setminus E_r) \). For a set of \( N \) randomly sampled points \( S = \{s_1, s_2, ..., s_N\} \subset V \), we compute the all-pairs shortest paths \( D_{GT} \) in the original network \( G \) and \( D_{pred} \) in the reconstructed network \( G^* \) produced by our model. The objective is to optimize \( G^* \) such that \( \sum D_{pred} \leq \sum D_{GT} \), ensuring the total shortest path length in \( G^* \) is minimized while restoring connectivity. The process continues iteratively until convergence.

\begin{figure}[h]
    \centering
    \includegraphics[width=\linewidth]{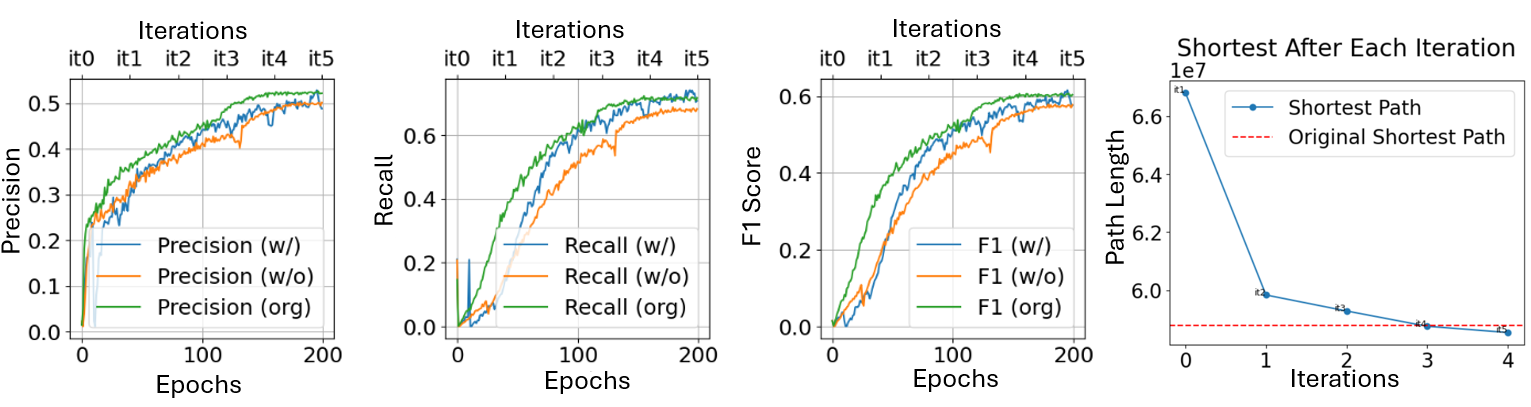}
    \caption{The plot illustrates how gaps in the ground truth (represented by the orange curve) impact model performance compared to the correct ground truth (shown by the green curve). The blue curve demonstrates improvement over time as the noisy ground truth is corrected using the IGraSS framework. The rightmost curve indicates how the shortest path length decreases as more breaks in the network are connected over time through the application of IGraSS.}
    \label{fig:road_results_main}
\end{figure}
\begin{table}[h]
\footnotesize
\centering
\begin{tabular}{l|r|r|r|r|r}
\toprule
\textbf{Models}                      & \textbf{Method} & \textbf{\rd} & \textbf{\rd} & \textbf{\rf} & \textbf{\rd} \\
\midrule
\multirow{3}{*}{ResNet50}   & Original            & 0.782                           & 0.852                           & 0.816      & 0.698                           \\
                                     & w/o                & 0.709                           & 0.834                           & 0.767       & 0.675                           \\
                                     & w                  & 0.752                           & 0.846                           & 0.797       & 0.691                           \\\midrule
\multirow{3}{*}{ResUnet}    & Original            & 0.859                           & 0.912                           & 0.885       & 0.732                           \\
                                     & w/o                & 0.825                           & 0.899                           & 0.859       & 0.699                           \\
                                     & w                  & 0.848                           & 0.910                            & 0.878       & 0.711                           \\\midrule
\multirow{3}{*}{DeepLabV3+} & Original            & 0.872                           & 0.943                           & 0.906       & 0.768                           \\
                                     & w/o                & 0.846                           & 0.929                           & 0.885       & 0.709                           \\
                                     & w                  & 0.861                           & 0.942                           & 0.898       & 0.737   \\
                                     \bottomrule
\end{tabular}
\caption{Performance Analysis on Road Datasets.}
\label{tab:road_performance}
\end{table}
\textbf{\noindent}{Results.}Our results show that IGraSS effectively reduces shortest path lengths to match the ground truth by the third iteration (see Fig.~\ref{fig:road_results_main}). Since the network remains well-connected despite small gaps, reachability is not a suitable constraint; instead, shortest path minimization yields better results. Performance metrics exhibit similar trends with and without the framework, as observed in the canal network. Comparing final outputs with the ground truth confirms that our approach achieves comparable results across all metrics (see Table~\ref{tab:road_performance}). Due to space constraints, extensive results are provided in the Appendix (Online Supplementary Materials).

\section{Discussion}
In summary, we developed a framework called IGraSS to address the challenges posed by weak and incomplete annotations by leveraging global constraints inherent in network-like infrastructures. We assessed IGraSS on canal networks with reachability constraints, successfully filling gaps in the ground truth and achieving improved results. IGraSS can be generalized to other network infrastructures with different constraints, as demonstrated with road networks. While we use DEM alongside RGB to enhance canal segmentation by providing elevation context, the refinement module currently relies solely on reachability. {We acknowledge that performance may degrade in areas where canals are obscured by vegetation or terrain, and future work will explore integrating flow direction and additional multi-modal data to improve robustness. By enabling accurate and automated mapping of infrastructure networks, IGraSS supports sustainable water and land resource management—advancing SDG 12 by promoting more responsible and efficient agricultural planning and infrastructure development.}

\section*{Ethical Statement}

There are no ethical issues.

\section*{Acknowledgments}
{This material is based upon work supported by the AI Research Institutes program supported by NSF and USDA-NIFA under the AI Institute: Agricultural AI for Transforming Workforce and Decision Support (AgAID) award No.~2021-67021-35344. This work was
partially supported by University of Virginia Strategic Investment Fund award number SIF160.}


\bibliographystyle{named}
\bibliography{ijcai25}

\begin{thebibliography}{}

\bibitem[\protect\citeauthoryear{Abdollahi \bgroup \em et al.\egroup }{2020}]{abdollahi2020deep}
Abolfazl Abdollahi, Biswajeet Pradhan, Nagesh Shukla, Subrata Chakraborty, and Abdullah Alamri.
\newblock Deep learning approaches applied to remote sensing datasets for road extraction: A state-of-the-art review.
\newblock {\em Remote Sensing}, 12(9):1444, 2020.

\bibitem[\protect\citeauthoryear{Archuleta and Terziotti}{2023}]{Archuleta_Terziotti_2023}
Christy-Ann~M. Archuleta and Silvia Terziotti.
\newblock Elevation-derived hydrography—representation, extraction, attribution, and delineation rules.
\newblock Technical report, U.S. Department of the Interior, U.S. Geological Survey, January 2023.

\bibitem[\protect\citeauthoryear{Bastani \bgroup \em et al.\egroup }{2018}]{bastani2018roadtracer}
Farzaneh Bastani, Songtao He, Miao Liu, Hamed Samet, and John Krumm.
\newblock Roadtracer: Automatic extraction of road networks from aerial images.
\newblock In {\em Proceedings of the IEEE Conference on Computer Vision and Pattern Recognition (CVPR)}, pages 4720--4728, 2018.

\bibitem[\protect\citeauthoryear{Belt and Smith}{2009}]{Belt2009}
Richard~L. Belt and Stephen~W. Smith.
\newblock {\em Infrastructure Inventory and GIS Mapping for Canal Irrigation Delivery Systems}.
\newblock U.S. Committee on Irrigation and Drainage, June 2009.
\newblock Presented at Irrigation District Sustainability - Strategies to Meet the Challenges: USCID Irrigation District Specialty Conference, June 3-6, 2009, Reno, Nevada.

\bibitem[\protect\citeauthoryear{Chen \bgroup \em et al.\egroup }{2018}]{deeplab}
Liang{-}Chieh Chen, Yukun Zhu, George Papandreou, Florian Schroff, and Hartwig Adam.
\newblock Encoder-decoder with atrous separable convolution for semantic image segmentation.
\newblock {\em CoRR}, abs/1802.02611, 2018.

\bibitem[\protect\citeauthoryear{Chen \bgroup \em et al.\egroup }{2022}]{chen2022ga}
Xin Chen, Qun Sun, Wenyue Guo, Chunping Qiu, and Anzhu Yu.
\newblock Ga-net: A geometry prior assisted neural network for road extraction.
\newblock {\em International Journal of Applied Earth Observation and Geoinformation}, 114:103004, 2022.

\bibitem[\protect\citeauthoryear{Cheng \bgroup \em et al.\egroup }{2021}]{cheng2021joint}
Mingfei Cheng, Kaili Zhao, Xuhong Guo, Yajing Xu, and Jun Guo.
\newblock Joint topology-preserving and feature-refinement network for curvilinear structure segmentation.
\newblock In {\em Proceedings of the IEEE/CVF International Conference on Computer Vision}, pages 7147--7156, 2021.

\bibitem[\protect\citeauthoryear{Cira \bgroup \em et al.\egroup }{2022}]{cira2022improving}
Calimanut-Ionut Cira, Martin Kada, Miguel-{\'A}ngel Manso-Callejo, Ram{\'o}n Alcarria, and Borja Bordel~Sanchez.
\newblock Improving road surface area extraction via semantic segmentation with conditional generative learning for deep inpainting operations.
\newblock {\em ISPRS International Journal of Geo-Information}, 11(1):43, 2022.

\bibitem[\protect\citeauthoryear{Creaco \bgroup \em et al.\egroup }{2023}]{creaco2023}
E.~Creaco, G.~Barbero, A.~Montanaro, et~al.
\newblock Effective optimization of irrigation networks with pressure-driven outflows at randomly selected installation nodes.
\newblock {\em Scientific Reports}, 13:19218, 2023.

\bibitem[\protect\citeauthoryear{{Demir~et~al.}}{2018}]{deepglobe2018}
{Demir~et~al.}
\newblock Deepglobe 2018: A challenge to parse the earth through satellite images.
\newblock In {\em Proceedings of the IEEE Conference on Computer Vision and Pattern Recognition Workshops (CVPRW)}, pages 172--181, 2018.

\bibitem[\protect\citeauthoryear{Diakogiannis \bgroup \em et al.\egroup }{2019}]{resunet}
Foivos~I. Diakogiannis, Fran{\c{c}}ois Waldner, Peter Caccetta, and Chen Wu.
\newblock Resunet-a: a deep learning framework for semantic segmentation of remotely sensed data.
\newblock {\em CoRR}, abs/1904.00592, 2019.

\bibitem[\protect\citeauthoryear{Dijkstra}{1959}]{dijkstra1959note}
Edsger~W. Dijkstra.
\newblock A note on two problems in connexion with graphs.
\newblock {\em Numerische Mathematik}, 1(1):269--271, 1959.

\bibitem[\protect\citeauthoryear{Fan \bgroup \em et al.\egroup }{2023}]{FAN2023107565}
Yu~Fan, Haorui Chen, Zhanyi Gao, and Xiaomin Chang.
\newblock Canal water distribution optimization model based on water supply conditions.
\newblock {\em Computers and Electronics in Agriculture}, 205:107565, 2023.

\bibitem[\protect\citeauthoryear{Fisher \bgroup \em et al.\egroup }{2003}]{Fisher2003}
Robert Fisher, Simon Perkins, Ashley Walker, and Erik Wolfart.
\newblock Hypermedia image processing reference (hipr2), 2003.

\bibitem[\protect\citeauthoryear{Ganaye \bgroup \em et al.\egroup }{2018}]{ganaye2018semi}
Pierre-Antoine Ganaye, Micha{\"e}l Sdika, and Hugues Benoit-Cattin.
\newblock Semi-supervised learning for segmentation under semantic constraint.
\newblock In {\em Medical Image Computing and Computer Assisted Intervention--MICCAI 2018: 21st International Conference, Granada, Spain, September 16-20, 2018, Proceedings, Part III 11}, pages 595--602. Springer, 2018.

\bibitem[\protect\citeauthoryear{Gharbia}{2023}]{gharbia2023deep}
Reham Gharbia.
\newblock Deep learning for automatic extraction of water bodies using satellite imagery.
\newblock {\em Journal of the Indian Society of Remote Sensing}, 51(7):1511--1521, 2023.

\bibitem[\protect\citeauthoryear{He \bgroup \em et al.\egroup }{2015}]{he2015deep}
Kaiming He, Xiangyu Zhang, Shaoqing Ren, and Jian Sun.
\newblock Deep residual learning for image recognition, 2015.

\bibitem[\protect\citeauthoryear{He \bgroup \em et al.\egroup }{2020}]{he2020sat2graph}
Songtao He, Favyen Bastani, Satvat Jagwani, Mohammad Alizadeh, Hari Balakrishnan, Sanjay Chawla, Mohamed~M Elshrif, Samuel Madden, and Mohammad~Amin Sadeghi.
\newblock Sat2graph: Road graph extraction through graph-tensor encoding.
\newblock In {\em Computer Vision--ECCV 2020: 16th European Conference, Glasgow, UK, August 23--28, 2020, Proceedings, Part XXIV 16}, pages 51--67. Springer, 2020.

\bibitem[\protect\citeauthoryear{He \bgroup \em et al.\egroup }{2022}]{he2022swin}
Xin He, Yong Zhou, Jiaqi Zhao, Di~Zhang, Rui Yao, and Yong Xue.
\newblock Swin transformer embedding unet for remote sensing image semantic segmentation.
\newblock {\em IEEE Transactions on Geoscience and Remote Sensing}, 60:1--15, 2022.

\bibitem[\protect\citeauthoryear{Hosseinzade \bgroup \em et al.\egroup }{2017}]{HOSSEINZADE2017177}
Zeinab Hosseinzade, Sheree~A. Pagsuyoin, Kumaraswamy Ponnambalam, and Mohammad~J. Monem.
\newblock Decision-making in irrigation networks: Selecting appropriate canal structures using multi-attribute decision analysis.
\newblock {\em Science of The Total Environment}, 601-602:177--185, 2017.

\bibitem[\protect\citeauthoryear{Li \bgroup \em et al.\egroup }{2022}]{li2022accurate}
Junjie Li, Yizhuo Meng, Yuanxi Li, Qian Cui, Xining Yang, Chongxin Tao, Zhe Wang, Linyi Li, and Wen Zhang.
\newblock Accurate water extraction using remote sensing imagery based on normalized difference water index and unsupervised deep learning.
\newblock {\em Journal of Hydrology}, 612:128202, 2022.

\bibitem[\protect\citeauthoryear{Liu \bgroup \em et al.\egroup }{2022}]{liu2022survey}
Pengfei Liu, Qing Wang, Gaochao Yang, Lu~Li, and Huan Zhang.
\newblock Survey of road extraction methods in remote sensing images based on deep learning.
\newblock {\em Journal of Photogrammetry, Remote Sensing and Geoinformation Science}, 90(2):135--159, 2022.

\bibitem[\protect\citeauthoryear{Loureiro \bgroup \em et al.\egroup }{2024}]{Loureiro2024}
D.~Loureiro, P.~Beceiro, E.~Fernandes, et~al.
\newblock Energy efficiency assessment in collective irrigation systems using water and energy balances: methodology and application.
\newblock {\em Irrigation Science}, 42:745--768, 2024.

\bibitem[\protect\citeauthoryear{Nagaraj and Kumar}{2024}]{nagaraj2024extraction}
R~Nagaraj and Lakshmi~Sutha Kumar.
\newblock Extraction of surface water bodies using optical remote sensing images: A review.
\newblock {\em Earth Science Informatics}, 17(2):893--956, 2024.

\bibitem[\protect\citeauthoryear{{NASA}}{2023}]{NASA_CSDAP}
{NASA}.
\newblock {Commercial Smallsat Data Acquisition Program}.
\newblock \url{https://earthdata.nasa.gov/esds/csdap}, 2023.
\newblock Accessed: 31 August, 2024.

\bibitem[\protect\citeauthoryear{{National Hydrography }}{2020}]{usgs_nhd}
{National Hydrography }.
\newblock {National Hydrography Dataset}.
\newblock \url{https://www.usgs.gov/national-hydrography/national-hydrography-dataset}, 2020.
\newblock U.S. Department of the Interior.

\bibitem[\protect\citeauthoryear{Nations}{2015}]{UN2015}
United Nations.
\newblock Transforming our world: The 2030 agenda for sustainable development, 2015.
\newblock Accessed: 2014-01-15.

\bibitem[\protect\citeauthoryear{P\'{e}rez-Blanco \bgroup \em et al.\egroup }{2020}]{Blanco2020}
C.~Dionisio P\'{e}rez-Blanco, Arthur Hrast-Essenfelder, and Chris Perry.
\newblock Irrigation technology and water conservation: A review of the theory and evidence.
\newblock {\em Review of Environmental Economics and Policy}, 14(2):216--239, 2020.

\bibitem[\protect\citeauthoryear{Ren \bgroup \em et al.\egroup }{2022}]{ren2022automated}
Simiao Ren, Wayne Hu, Kyle Bradbury, Dylan Harrison-Atlas, Laura~Malaguzzi Valeri, Brian Murray, and Jordan~M Malof.
\newblock Automated extraction of energy systems information from remotely sensed data: A review and analysis.
\newblock {\em Applied Energy}, 326:119876, 2022.

\bibitem[\protect\citeauthoryear{{SpaceNet on Amazon Web Services (AWS)}}{2018}]{spacenet}
{SpaceNet on Amazon Web Services (AWS)}.
\newblock {Datasets}.
\newblock \url{https://spacenet.ai/datasets/}, 2018.
\newblock The SpaceNet Catalog. Last modified October 1st, 2018. Accessed on Sep 2, 2024.

\bibitem[\protect\citeauthoryear{{USGS}}{2024}]{USGS_3DEP}
{USGS}.
\newblock U.s. geological survey 3d elevation program (3dep), 1-meter digital elevation model (dem), 2024.

\bibitem[\protect\citeauthoryear{Xu}{2006}]{xu2006modification}
Hanqiu Xu.
\newblock Modification of normalised difference water index (ndwi) to enhance open water features in remotely sensed imagery.
\newblock {\em International journal of remote sensing}, 27(14):3025--3033, 2006.

\bibitem[\protect\citeauthoryear{Yu \bgroup \em et al.\egroup }{2023}]{yu2023boundary}
Jie Yu, Yang Cai, Xin Lyu, Zhennan Xu, Xinyuan Wang, Yiwei Fang, Wenxuan Jiang, and Xin Li.
\newblock Boundary-guided semantic context network for water body extraction from remote sensing images.
\newblock {\em Remote Sensing}, 15(17):4325, 2023.

\bibitem[\protect\citeauthoryear{Zhang and Long}{2023}]{dualGraph}
Liang Zhang and Cheng Long.
\newblock Road network representation learning: A dual graph-based approach.
\newblock {\em ACM Trans. Knowl. Discov. Data}, 17(9), jun 2023.

\bibitem[\protect\citeauthoryear{Zhou \bgroup \em et al.\egroup }{2018}]{Zhou2018DLinkNetLW}
Lichen Zhou, Chuang Zhang, and Ming Wu.
\newblock D-linknet: Linknet with pretrained encoder and dilated convolution for high resolution satellite imagery road extraction.
\newblock {\em 2018 IEEE/CVF Conference on Computer Vision and Pattern Recognition Workshops (CVPRW)}, pages 192--1924, 2018.

\end{thebibliography}
\clearpage
\appendix
\setcounter{figure}{0}
\setcounter{table}{0}
\renewcommand{\thefigure}{S\arabic{figure}}
\renewcommand{\thetable}{S\arabic{table}}

\twocolumn[

\begin{center}
\parbox{\textwidth}{\centering\Large\textbf{Supplementary Information: \\
IGraSS: Learning to Identify Infrastructure Networks from Satellite Imagery by
Iterative Graph-constrained Semantic Segmentation}}
\end{center}
\bigskip
] 

\bigskip
\baselineskip = 1.10\normalbaselineskip
\section{Dataset}
\label{sec:Dataset}

\begin{figure}[h]
    \centering
    \begin{subfigure}[b]{0.32\textwidth}
        \centering
        \includegraphics[width=\textwidth]{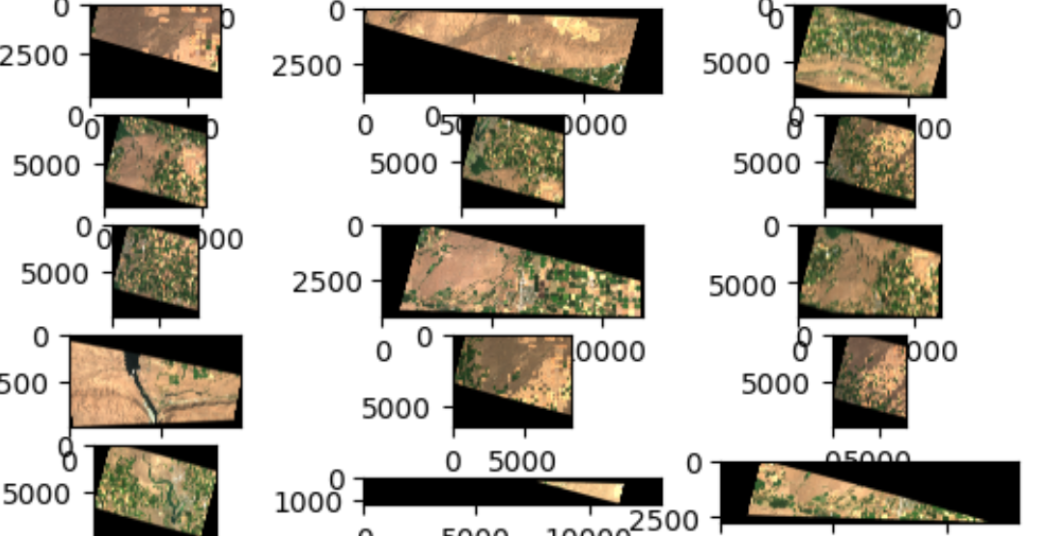}
        \caption{Image Tiles in Dataset}
        \label{fig:data-patch}
    \end{subfigure}
    \hfill
    \begin{subfigure}[b]{0.32\textwidth}
        \centering
        \includegraphics[width=\textwidth]{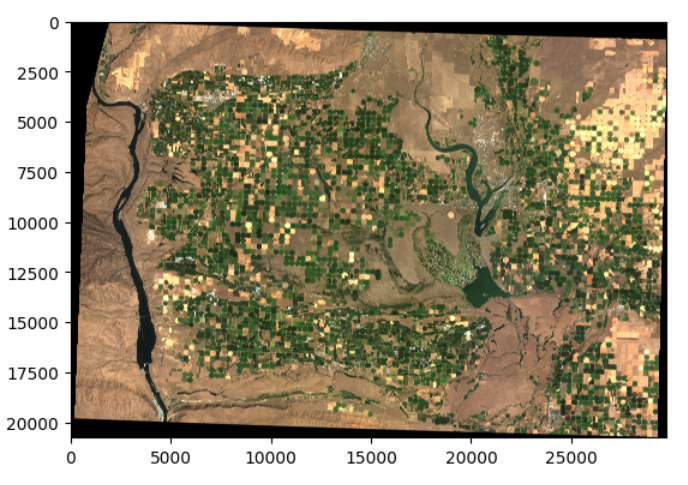}
        \caption{Merged Large Image Tile}
        \label{fig:large-patch}
    \end{subfigure}
    \hfill
    \begin{subfigure}[b]{0.32\textwidth}
        \centering
        \includegraphics[width=\textwidth]{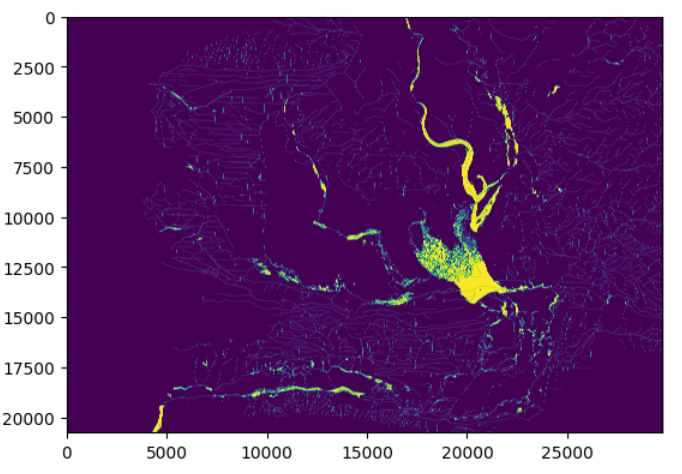}
        \caption{Corresponding Mask}
        \label{fig:mask}
    \end{subfigure}
    \caption{Visualization of Dataset}
    \label{fig:all-images}
\end{figure}

PlanetScope is a constellation of 180+ CubeSats that began imaging the Earth's surface in 2016. In this study, we used PlanetScope Ortho Tile. PlanetScope Ortho tiles are available in 25 x 25 km tiles with a spatial resolution of 3.125 meters. Three bands were used: blue (440--510 nm), green (520--590 nm) and red (630--685 nm) for map irrigation canal infrastructure network.
PlanetScope optical satellite imagery from 2020 to 2023 consists of approximately 30-35 non-identical shape tiles for each year (fig. \ref{fig:data-patch}).  Image selection was based on availability, cloud cover, and spatial coverage, provided through the Commercial Smallsat Data Acquisition Program. We first merge the tiles of each year based on their geographic locations, generating a comprehensive dataset for the entire Washington region with dimensions of $(20722, 29626, 3)$ for each year (See fig \ref{fig:large-patch} and \ref{fig:mask}). We mapped the NHD shapefile data onto our 
satellite imagery to generate masks. Subsequently, we create $512 \times 512$ pixel patches from these large patches to form our training dataset. In the process of generating the training set, we apply two filtering criteria: we exclude image patches that contain more than 30\% black pixels, and we omit mask patches that have less than 0.5\% canal pixels.

\textbf{Dataset Split.} For our main experiment, we generated two distinct sets (Set $1$ and Set $2$) of training, validation, and test data, all spatially separated. To create the test set, we divided the larger patches into two parts: 80\% for training and validation, and 20\% for testing. This process was repeated to produce the second set of training, validation, and test data.

We do not modify or examine any data from the test set while running IGraSS. For the training and validation sets, we use 5-fold cross-validation and run IGraSS to track the average output of the metrics. After finalizing the best parameters (iterations, epochs, $\rho$ etc.), we run IGraSS again on the training sets and evaluate the performance on the corresponding test sets.

\begin{figure}
    \centering
    \includegraphics[width=\linewidth]{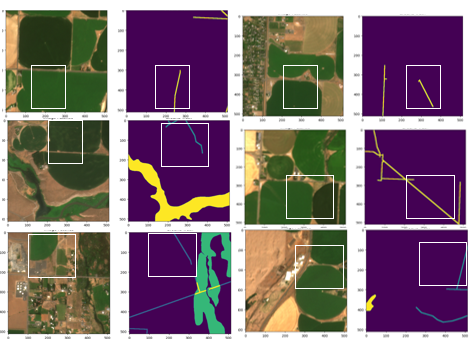}
    \caption{Hard to predict Images}
    \label{fig:hardpredictions}
\end{figure}

\section{Additional Framework Implementation Details For Canal Network}
\label{sec:ap_framework}
In this section, we will discuss each component of the IGraSS framework in detail.
\begin{figure*}
    \centering
    \includegraphics[width=\linewidth]{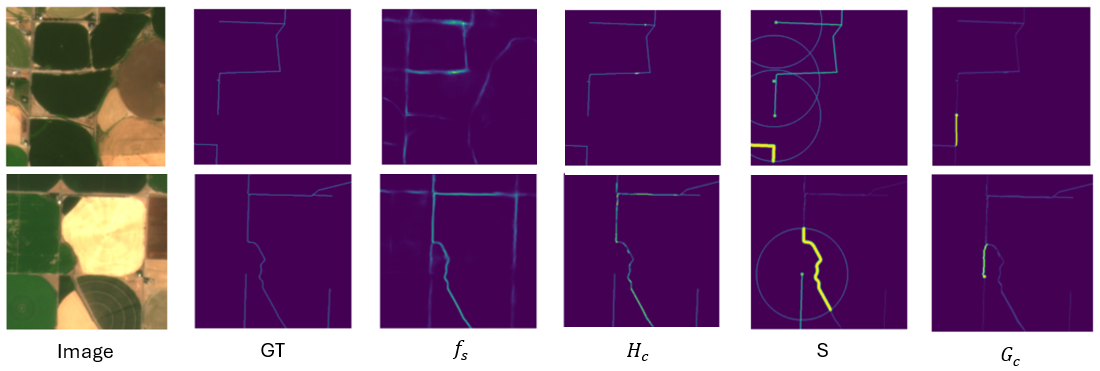}
    \caption{Caption}
    \label{fig:frameworkViz}
\end{figure*}
\paragraph{Learner:}
We take the trained learner and make predictions on the train set and generate $\fs^i(v)$.

\paragraph{Pre-completion network:} Given the learner's output, we use a threshold $\tau = 0.5$ to compute the likelihood $w^i(v)$. Since the width of the output prediction of canal segments can be wider than 1 pixel, we initially perform a morphological dilation on the $\cg^i$ with a kernel size of 5 and merge $\cg^i$ with the result of $w^i(v) > \tau$. We then perform morphological erosion to obtain canal segments with a width of one pixel, resulting in the pre-completion network $\pcg^i$.(See fig \ref{fig:frameworkViz})

\subsection{NDWI}
Difference Water Index (NDWI) are commonly used indices in remote sensing. NDWI, on the other hand, is used for detecting water bodies and assessing vegetation water content, calculated as \( NDWI = \frac{(GREEN - NIR)}{(GREEN + NIR)} \), where GREEN represents the green reflectance.
\paragraph{Network-completion instances.} For \emph{network completion instances} ~$\sti$ consists of a tuple~$(t,S_t,H_t)$, we compute the \emph{reachable} $\labr(v)=1$ and \emph{unreachable nodes} $\labr(v)=0$, \emph{unreachable terminal}~$t$,~$S_t=\{s_1,s_2,\ldots\}$
is a collection of reachable nodes called \emph{sources}, and $H_t$, 
referred to as the \emph{$t$-local graph}, is a subgraph of the grid
graph~$G$ containing~$\{t\}\cup S_t$.

\textbf{Reachable and Unreachable Nodes}. Let the water source indices be defined as the set $F = \{(i_1, j_1), \ldots, (i_k, j_k)\}$. We begin by determining the canal pixels directly connected to these water sources using Algorithm~\ref{alg:directly_connected}. The Directly Connected Canals algorithm operates on a binary matrix $M \in \{0,1\}^{m \times n}$ where $M$ represents the canal network (here pre-completion $\pcg^i$) and a set of initial index pairs $F = \{(i_1, j_1), \ldots, (i_k, j_k)\}$. It constructs a boolean mask $B \in \{0,1\}^{m \times n}$ where $B_{i,j} = 1$ if $(i,j) \in F$, and 0 otherwise. The algorithm then defines an 8-connectivity kernel $K = \begin{bmatrix} 1 & 1 & 1 \\ 1 & 0 & 1 \\ 1 & 1 & 1 \end{bmatrix}$ and computes the convolution $V = B * K$, where $*$ denotes 2D convolution with `same' mode and zero padding. Subsequently, it constructs a set $C = \{(i,j) \mid V_{i,j} > 0 \land M_{i,j} = 1, 1 \leq i \leq m, 1 \leq j \leq n\}$, representing the directly connected 1s. The algorithm's output consists of the set $C$ of directly connected canals(1s). These pixels in $C$ are immediately classified as reachable. Subsequently, we employ a breadth-first search (BFS) algorithm, initializing it with each pixel $p \in C$. The BFS traverses through all connected pixels, marking each visited pixel as reachable. Formally, let $R$ be the set of all reachable pixels. We define $R$ as:
\[
R = C \cup \bigcup_{p \in C} \text{BFS}(M, \{p\}, C)
\]
where $M$ is the binary matrix representing the canal network and $\text{BFS}(M, \{p\}, C)$ returns the set of all pixels reachable from $p$ through connected canal pixels. Once we have the reachable canal pixels, we simply identify the rest of the canal pixels as non-reachable canal pixel.  We define the set of non-reachable canal pixels, denoted as $U$,  -- the set of ones in \( M \) excluding the indices present in \( R \) is given by:
\[
U' = \{ (i, j) \mid M[i, j] = 1 \text{ and } (i, j) \notin R \}
\]
In the subsequent step, we compute the set of unreachable nodes $U_G$ from the ground truth network $\gtg$, utilizing the same process described above. We then refine our set of unreachable nodes $U$ by intersecting it with $U_{\gtg}$:
\[
U = U' \cap \gtg
\]
This refinement ensures that we retain only those unreachable nodes in $U$ that are also present in the set of unreachable nodes of the ground truth network $\gtg$.

%
\textbf{Terminals.} Given the set of unreachable nodes ~$U$, now we identify the terminal nodes. The Terminal Nodes Detection Algorithm  \ref{alg:terminals} identifies terminals in a unreachable canal network represented as a set of coordinates $U$. For each point $p \in U$, the algorithm examines its 8-connected neighborhood defined by the directions $\Delta = \{(0, \pm1), (\pm1, 0), (\pm1, \pm1)\}$. A point is classified as a terminal if it has one or fewer neighbors within the set $U$. The algorithm maintains a set $V$ of visited points to avoid redundant computations. Finally, it returns the set of terminals $E$.

\textbf{Source Terminal Pairs.} For each terminal point $t \in E$, we identify its corresponding source points $S_t$ within a user-specified radius $\rho$. The source set of $t$ comprises all points in the local graph $H_t$ that are reachable. These source points can be either water sources or reachable canal nodes. From previously calculated water sources, we first identify their edge points using an algorithm similar to Algorithm~\ref{alg:terminals}. However, in this case, we designate as edge points those which have fewer than eight neighbors. To find the sources $S_t$ for each terminal $t$, we begin by aggregating all potential source points, including both the identified water source edges and the reachable canal nodes. Next, we use Algorithm \ref{alg:source} to identify source terminal pairs.

\begin{algorithm}[h]
\caption{Identifying Source-Terminal Pairs}
\KwIn{$T$: Terminal points \\
      $S$: Set of potential source points (water sources and reachable canal nodes) \\
      $\rho$: User-specified radius}
\KwOut{$S_t$: Set of source-terminal pairs}

Compute distances as \( \sqrt{\sum_{i=1}^{d} (S_t[i] - t)^2} \)\;
$S_t \gets \{p \in S \mid \text{distance}(p, t) \leq \rho\}$\;

\Return $S_t$\;
\label{alg:source}
\end{algorithm}
Next, we process the nodes in \emph{$t$-local graph} $H_t$, 
which is a subgraph of the grid
graph~$\cg^i$ containing~$\{t\}\cup S_t$ where each node~$v\in V(H_t)$ has 
weight~$1/w^i(v)$.
Given a set of endpoints $E$, likelihood~$w^i(v)$ matrix, the pre-completion network $\pcg^i$, radius $\rho$, threshold $th$ -- referred as confidence threshold $\alpha$ algorithm \ref{alg:edge} processes each terminal point $t \in E$. For each $t$, it computes the set of neighbors $N_p$ within a radius $\rho$. Each neighbor $n \in N_p$ is evaluated: if its weight $w^i[n]$ exceeds a threshold $th$ and $\pcg^i[n] = 0$, it's added to the result matrix with a value inversely proportional to its weight ~$1/w^i(v)$. The algorithm the resultant matrix $X_r$. 
\begin{algorithm}[h]
\caption{Terminal Nodes Detection}
\KwIn{$U$: Set of unreachable canal pixel coordinates}
\KwOut{$E$: Set of endpoint coordinates \\
       $V$: Set of visited coordinates}

$V \gets \emptyset$\;
$E \gets \emptyset$\;
$\Delta \gets \{(0, \pm1), (\pm1, 0), (\pm1, \pm1)\}$\;

\ForEach{$p \in U$}{
    \If{$p \notin V$}{
        $N \gets \emptyset$\;
        \ForEach{$\delta \in \Delta$}{
            $n \gets p + \delta$\;
            \If{$n \in U$}{
                $N \gets N \cup \{n\}$\;
            }
        }
        \If{$|N| \leq 1$}{
            $E \gets E \cup \{p\}$\;
        }
        $V \gets V \cup \{p\}$\;
    }
}

\Return $E$\;
\label{alg:terminals}
\end{algorithm}
\paragraph{Network reachability computation.} Finally, we compute the set of \emph{network completion instances} $\stic$. Each instance $\sti = (t, S_t, H_t) \in \stic$ consists of a terminal point $t$, its corresponding set of source points $S_t$, and the local graph $H_t$ generated from $X_r$ (See Algorithm \ref{alg:subgraph}). The algorithm operates on a matrix $M_{m\times n}$ representing a weighted grid (here we pass $X_r$), given a terminal point $t=(t_x,t_y)$ and a radius $\rho$. Initially, it constructs a subgraph $H_t=(V_H,E_H)$ within the region $R={(i,j) : |i-t_x| \leq \rho, |j-t_y| \leq \rho, 0 \leq i < m, 0 \leq j < n}$. The algorithm includes $(i,j)$ in $V_H$ for each point $(i,j) \in R$ where $M_{i,j} > 0$. Edges in $E_H$ are established between neighboring points in the 8-directional neighborhood $D={(\pm1,0), (0,\pm1), (\pm1,\pm1)}$. Subsequently, the algorithm reduces this node-weighted subgraph $H_t$ to an edge-weighted directed graph $G'=(V',E')$ using Folklore Algorithm. For each node $(i,j) \in V_H$, the algorithm creates two nodes $(i,j,1)$ and $(i,j,2)$ in $V'$, and adds a directed edge $((i,j,1),(i,j,2))$ to $E'$ with weight $w'(((i,j,1),(i,j,2))) = M_{i,j}$. For each edge between neighboring points $(i,j)$ and $(i+d_x,j+d_y)$ in $H$, where $(d_x,d_y) \in \Delta$, the algorithm adds directed edges $((i,j,2),(i+d_x,j+d_y,1))$ and $((i+d_x,j+d_y,2),(i,j,1))$ to $E'$, both with weight zero. This reduction preserves path weights, enabling any shortest path algorithm applied to $G'$ to solve the corresponding node-weighted shortest path problem in the original subgraph $H_t$, effectively determining optimal paths within the specified radius around the terminal point in the original grid.
The objective for each instance is to find the minimum weighted shortest path from $t$ to any point in $S_t$ within $H_t$. For each terminal $t$, we select the shortest path among all $(s, t)$ pairs, where $s \in S_t$. We then update $\pcg^i$ by marking the points along this path as canal pixels. This process generates the next iteration of the canal graph, denoted as $\cg^{i+1}$.

\section{Additional Experiments}
\label{sec:experiments}
\begin{figure*}[ht]
    \centering
    \begin{subfigure}[b]{0.48\textwidth}
        \includegraphics[width=\textwidth]{figs/canal.PNG}
        \caption{Visualization of Canal Network Completion:
        Blue lines represent reachable canal pixels, while red lines indicate unreachable canal pixels. The images demonstrate gaps in the red canal segments that are iteratively filled. Initially, the green segments connect one of the unreachable red segments with the blue reachable segment, making the upper red segment reachable in Iteration 1 (It~\#1). In the next iteration (It~\#2), yellow segments connect the smaller unreachable red segments. Finally (It~\#3), pink segments connect the remaining bottom segment by filling the gaps, thus making these canals reachable.}
        \label{fig:canal_viz}
    \end{subfigure}
    \hfill
    \begin{subfigure}[b]{0.45\textwidth}
        \includegraphics[width=\textwidth]{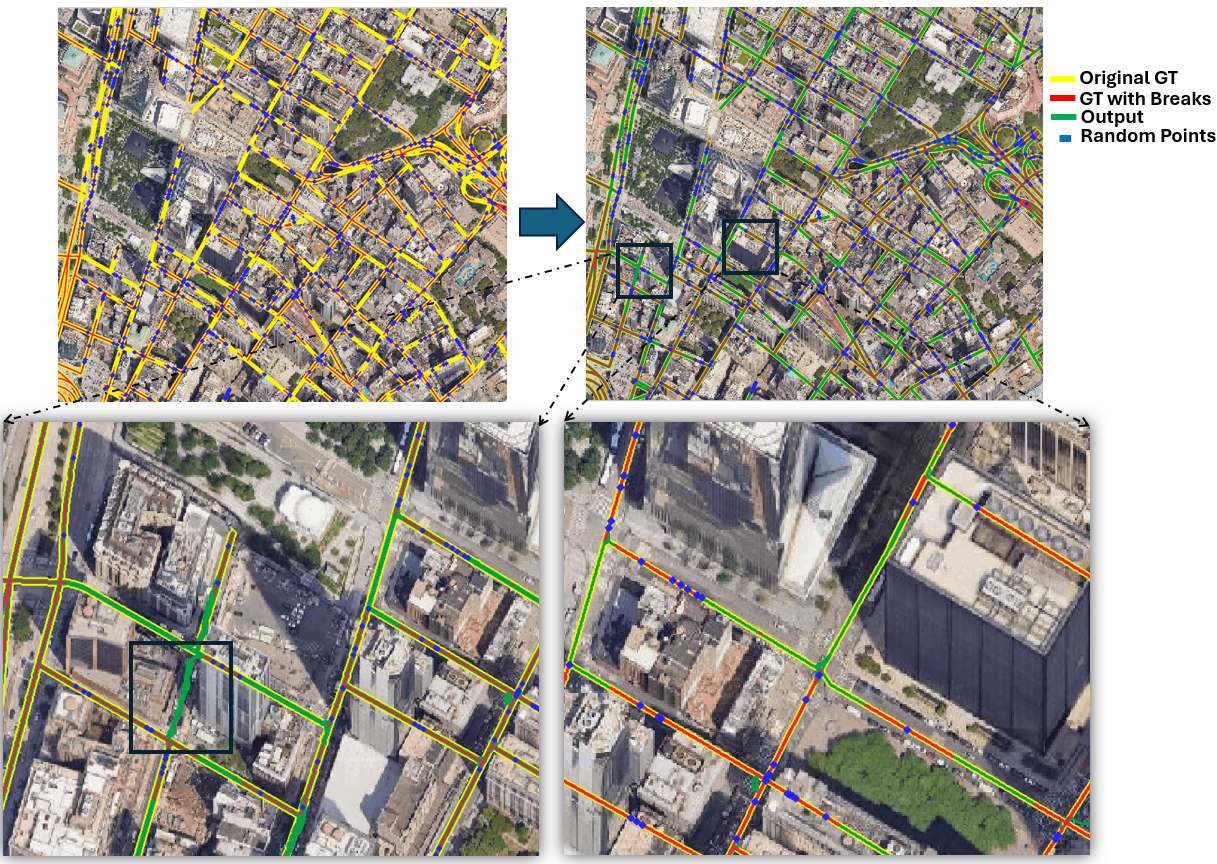}
        \caption{Visualization of Road Network Completion:
        The yellow lines represent the main ground truth of the road network. Red lines overlaying the yellow lines visualize randomly created breaks in the network. Blue points indicate randomly selected locations used to calculate the shortest paths. The green lines, in conjunction with the red lines, demonstrate how the breaks have been filled using the IGraSS framework. The zoomed-in image on the left shows that IGraSS has connected a new segment not present in the main ground truth, which happens to be an actual road.}
        \label{fig:road_network}
    \end{subfigure}
    \caption{Comparison of Canal and Road Network Completion using IGraSS}
    \label{fig:network_completion_comparison}
\end{figure*}
\begin{figure}[!h]
    \centering
    \includegraphics[width=\linewidth]{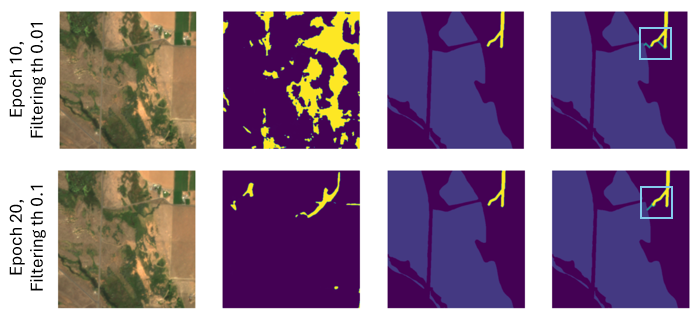}
    \caption{Error Results}
    \label{fig:error_results}
\end{figure}
\begin{figure*}
  
        \centering
        \includegraphics[width=\textwidth]{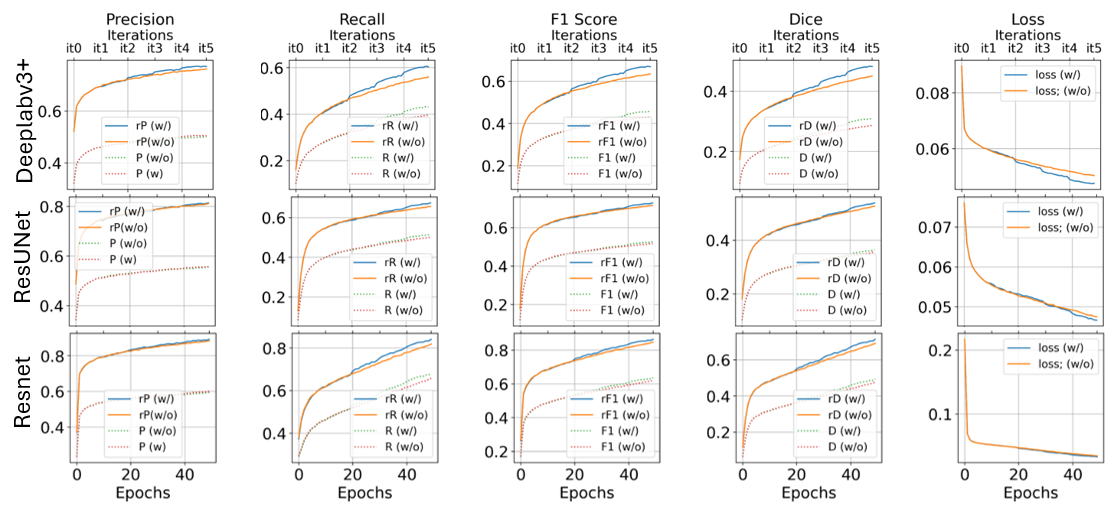}
        \caption{Performance comparison of models with and without IGraSS framework across epochs on train set 1}
        \label{fig:baseline_epoch1}
\end{figure*}
\begin{figure*}
  
        \centering
        \includegraphics[width=\textwidth]{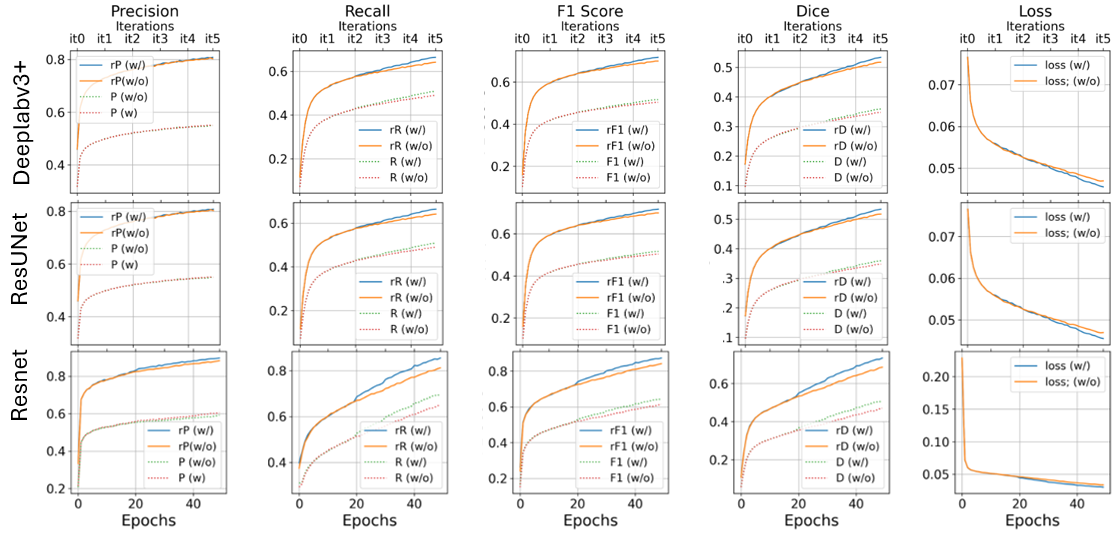}
        \caption{Performance comparison of various models with and without IGraSS framework across epochs on train sets 2}
        \label{fig:baseline_epoch12}
\end{figure*}

\begin{table}[h]
\footnotesize
\caption{Parameter Analysis with different radius $\rho$}
\begin{tabular}{c|c|ccc}
\toprule
Model                       & \multicolumn{1}{l}{$\rho$} & \multicolumn{1}{l}{Reachable} & \multicolumn{1}{l}{Unreachable} & \multicolumn{1}{l}{Terminals} \\
\midrule
\multirow{4}{*}{deeplabv3+} & 150                    & 836350                                      & 25163                                              & 124                                \\
                            & 100                    & 823965                                      & 37647                                              & 166                                \\
                            & 50                     & 821695                                      & 37885                                              & 178                                \\
                            & 20                     & 815048                                      & 44375                                              & 198                                \\\midrule
\multirow{4}{*}{resnet}     & 150                    & 842171                                      & 21662                                              & 97                                 \\
                            & 100                    & 831639                                      & 28732                                              & 133                                \\
                            & 50                     & 817908                                      & 41642                                              & 192                                \\
                            & 20                     & 814650                                      & 44742                                              & 202                                \\\midrule
\multirow{4}{*}{resunet}    & 150                    & 829062                                      & 32168                                              & 150                                \\
                            & 100                    & 819115                      &  40973                           & 189            \\
                            & 20                     & 814566                                      & 44843                                              & 204                                \\
                            & 50                     & 816974                                      & 42458                                              & 204                 \\
                            \bottomrule
\end{tabular}
\label{tab:parameter_tune}
\end{table}
\paragraph{Experimental Evaluation.}
\textbf{Training Results. } Figure \ref{fig:baseline_epoch1} and \ref{fig:baseline_epoch12} presents the training time results on the training sets, comparing the baseline model performance with the IGraSS framework using the same baseline as the Learner. Both approaches were trained for an equal number of epochs on both test sets. These models were trained using parameters $\rho = 100$, $th=0.1$, with 20 epochs per iteration for 5 iterations. Examining the precision, we observe minimal improvement over time. This can be attributed to the model learning to predict gaps more accurately, which paradoxically reduces precision when compared to the ground truth, as these newly predicted segments are not present in the original data. However, we note improvements in recall, F1 score, and Dice coefficient in both training set results. It's important to highlight that training set two has fewer unreachable canals than training set one, resulting in a lower impact of the IGraSS framework. This difference in impact demonstrates that IGraSS does not lead to connecting unnecessary gaps, but rather focuses on improving connectivity where needed.

\begin{figure*}[h]
    \centering
    \includegraphics[width=\textwidth]{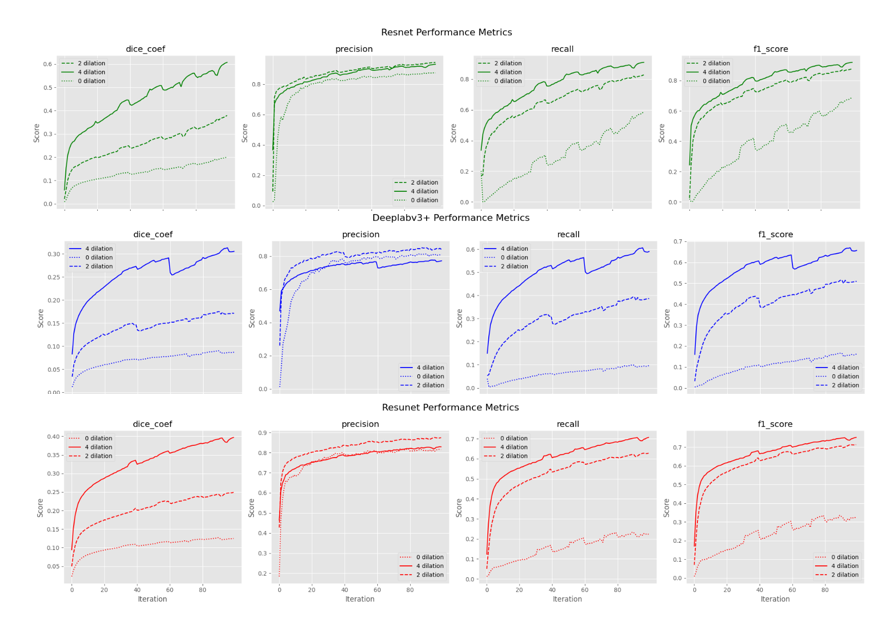}
    \caption{Training results various models using dilated convolutions with kernel sizes of 2, 4, and 8 applied to the ground truth mask.}
    \label{fig:dilaion}
\end{figure*}

\begin{figure*}
    \centering
    \begin{subfigure}[b]{0.48\textwidth}
        \centering
        \includegraphics[width=\textwidth]{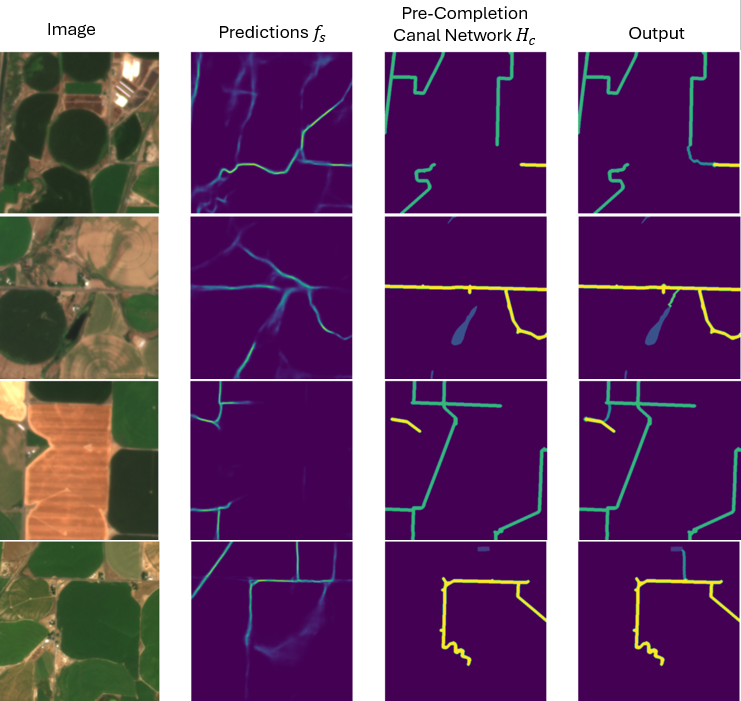}
        \label{fig:main-result1}
    \end{subfigure}
    \hfill
    \begin{subfigure}[b]{0.48\textwidth}
        \centering
        \includegraphics[width=\textwidth]{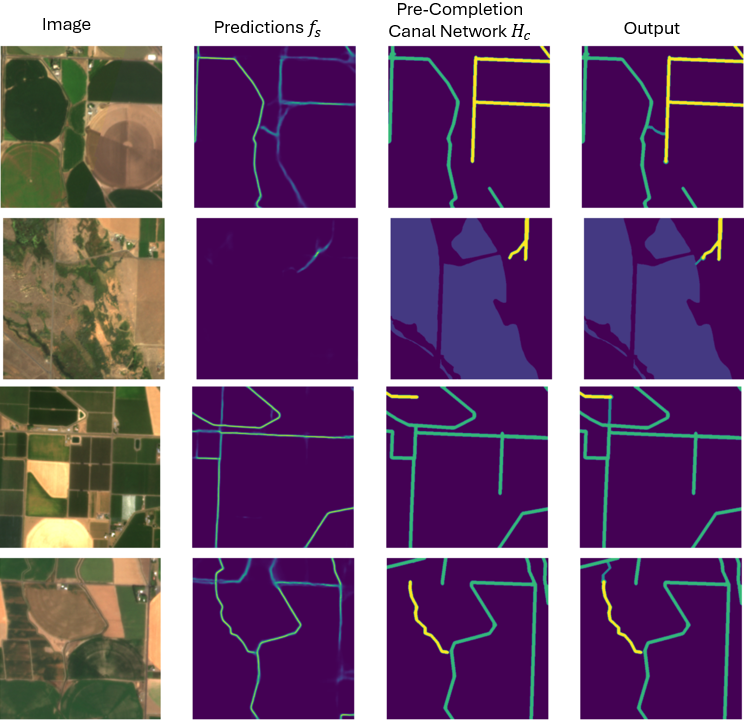}
        \label{fig:main-result2}
    \end{subfigure}
    \caption{Results of the network reachability analysis}
    \label{fig:main-results}
\end{figure*}

\textbf{Rechability Analysis on Test Set.} Table \ref{tab:canal_pixel_detection} presents a reachability analysis on both test sets, comparing the performance of the IGraSS framework with the baseline model. The training setup was consistent with the previously mentioned parameters. Results indicate that IGraSS outperforms the baseline in predicting reachable canals and overall total pixels in the ground truth. Regarding unreachable canals, IGraSS predicted a higher number than the ground truth, which can be attributed to its overall higher pixel prediction rate. Errors in reachability assessment may occur due to ground truth inconsistencies, such as underground canals or those that have changed over time but remain in the ground truth data (see Figure \ref{fig:hardpredictions}). In numerous images, canals present in the ground truth or connected to water sources are not visible, making it impossible for the model to predict them accurately. When the model correctly predicts the rest of the visible canals, the reachability performance may suffer due to the loss of these non-visible, yet critical connecting segments. Nevertheless, our results demonstrate significant improvement in overall performance, particularly in predicting a higher number of canals present in the ground truth. Fig \ref{fig:main-results} visualizes results on the test set 2 of with IGraSS.
\begin{figure*}[h]
    \centering
    \includegraphics[width=\linewidth]{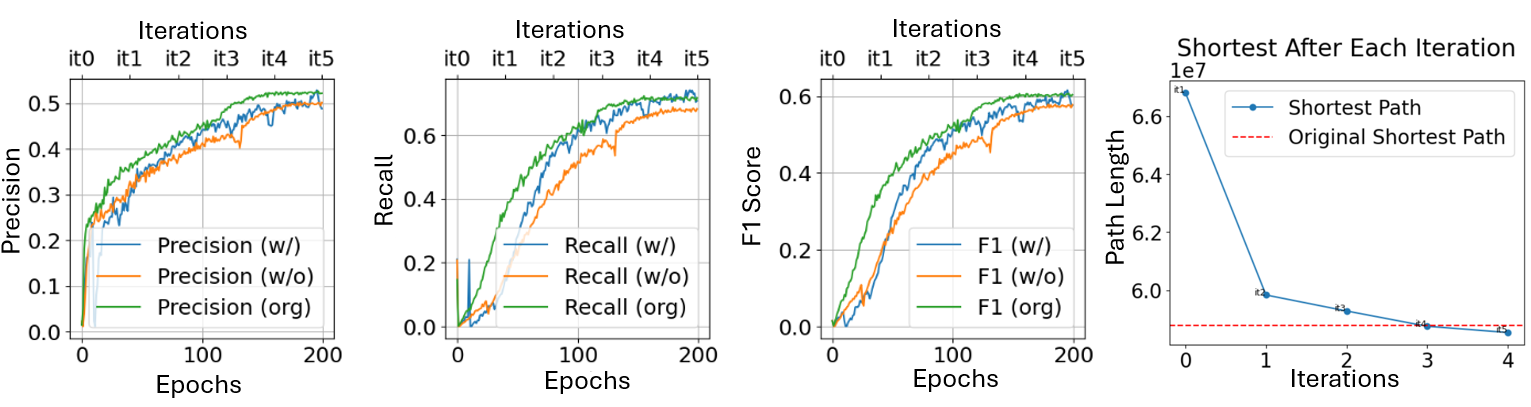}
    \caption{The plot illustrates how gaps in the ground truth (represented by the orange curve) impact model performance compared to the correct ground truth (shown by the green curve). The blue curve demonstrates improvement over time as the noisy ground truth is corrected using the IGraSS framework. The rightmost curve indicates how the shortest path length decreases as more breaks in the network are connected over time through the application of IGraSS.}
    \label{fig:road_results}
\end{figure*}
\paragraph{Dilation}
Given that our ground truth consists of one-pixel-wide lines, we conducted additional experiments to investigate whether dilation improves the model's learning performance. Intuitively, canals can vary in width, and predicting a single pixel line for canals of different sizes may challenge the model's learning capacity, potentially leading to slight location discrepancies in predictions. We hypothesized that increasing the width of the ground truth mask would enhance the model's learning ability. Our results in  Fig. \ref{fig:dilaion} demonstrate improvement across all metrics with a dilation of 2 pixels. Further dilation to 4 pixels shows additional improvements in the Dice coefficient and overall F1 score, although precision gains are limited across the models. However, we selected a dilation level of 4 for our subsequent analysis, as it demonstrated significant improvements in overall F1 score and Dice coefficient without substantially compromising precision.

\begin{table*}[h]
\centering
\begin{tabular}{l|c|l|r|r|r|r}
\toprule
\textbf{Test\_Set} & \textbf{Model} & \textbf{\begin{tabular}[c]{@{}l@{}}Output \\ Type\end{tabular}} & \textbf{\begin{tabular}[c]{@{}r@{}}Reachable \\ Canal\end{tabular}} & \textbf{\begin{tabular}[c]{@{}r@{}}Unreachable \\ Canal\end{tabular}} & \textbf{\begin{tabular}[c]{@{}l@{}}Total common \\ Canal pixel ($\tau=0.5$)\end{tabular}} & \textbf{\begin{tabular}[c]{@{}l@{}}Total common \\ Canal pixel \%\end{tabular}} \\ \midrule
 & & \textbf{GT} & \textbf{180,897} & \textbf{112,515} & \textbf{293,412} & \\ \cline{2-7} 
 & & IGraSS & 78,738 & 71,524 & 150,262 & 52\% \\ \cline{3-7} 
 & \multirow{-2}{*}{DeepLabv3+} & Baseline & 75,492 & 64,792 & 140,284 & 48\% \\ \cline{2-7} 
 & & IGraSS & 58,750 & 115,527 & 174,277 & 61\% \\ \cline{3-7} 
 & \multirow{-2}{*}{Resunet} & Baseline & 53,154 & 115,790 & 168,944 & 57\% \\ \cline{2-7} 
 & & IGraSS & 61,196 & 152,994 & 214,190 & 73\% \\ \cline{3-7} 
\multirow{-7}{*}{Set 1} & \multirow{-2}{*}{Resnet} & Baseline & 50,462 & 139,255 & 189,717 & 64\% \\ \midrule
 & & \textbf{GT} & \textbf{125,922} & \textbf{90,598} & \textbf{216,520} & \\ \cline{2-7} 
 & & IGraSS & 59,877 & 33,486 & 93,363 & 43\% \\ \cline{3-7} 
 & \multirow{-2}{*}{DeepLabv3+} & Baseline & 32,492 & 58,819 & 91,311 & 42\% \\ \cline{2-7} 
 & & IGraSS & 49,256 & 58,178 & 107,434 & 48\% \\ \cline{3-7} 
 & \multirow{-2}{*}{Resunet} & Baseline & 45,209 & 52,529 & 97,738 & 45\% \\ \cline{2-7} 
 & & IGraSS & 72,556 & 62,368 & 134,924 & 62\% \\ \cline{3-7} 
\multirow{-7}{*}{Set 2} & \multirow{-2}{*}{Resnet} & Baseline & 67,254 & 64,798 & 132,052 & 61\% \\ \bottomrule
\end{tabular}
\caption{Comparison of Canal Pixel Detection by Baseline Models and IGraSS Framework on both test sets.}
\label{tab:canal_pixel_detection}
\end{table*}


\begin{figure*}
    \centering
        \begin{subfigure}[b]{0.48\textwidth}
        \centering
        \includegraphics[width=\textwidth]{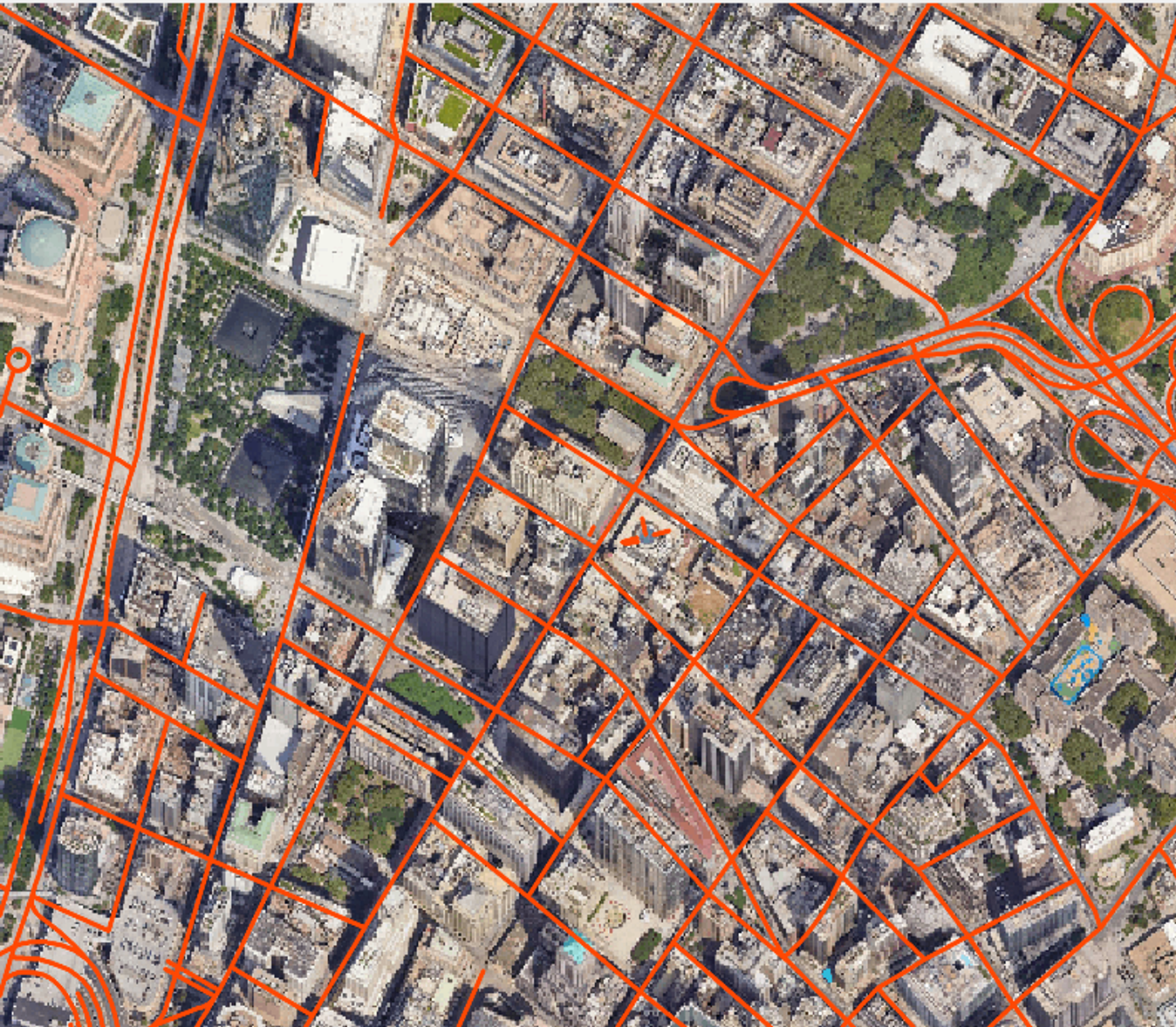}
        \caption{Ground Truth Network of Smaller Region of New York City}
        \label{fig:baseline_epoch2}
    \end{subfigure}
    \hfill
    \begin{subfigure}[b]{0.48\textwidth}
        \centering
        \includegraphics[width=\textwidth]{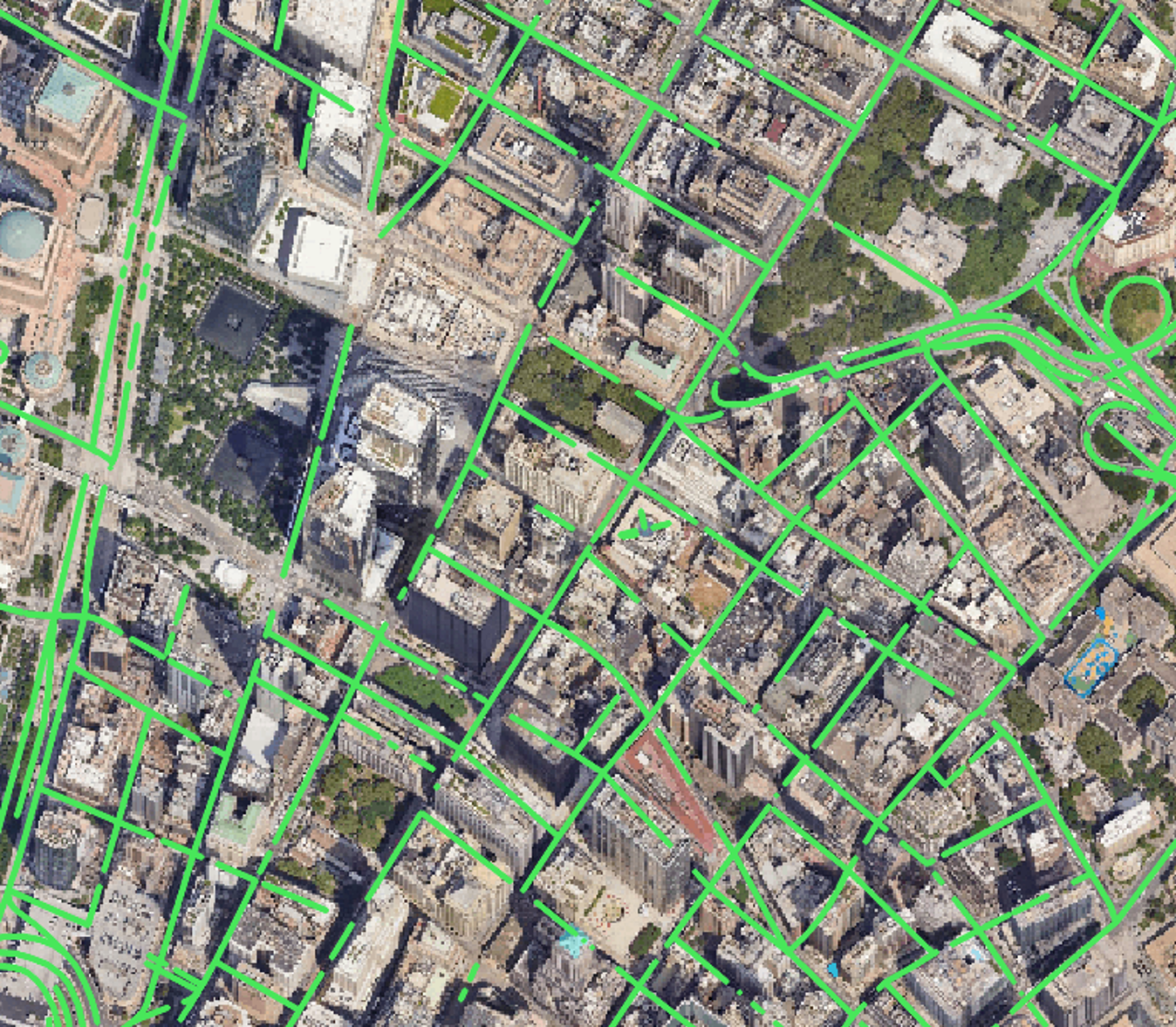}
        \caption{Randomly Induced Discontinuities in the Ground Truth Data}
        \label{fig:baseline_epoch2}
    \end{subfigure}
    \hfill
    \begin{subfigure}[b]{0.48\textwidth}
        \centering
        \includegraphics[width=\textwidth]{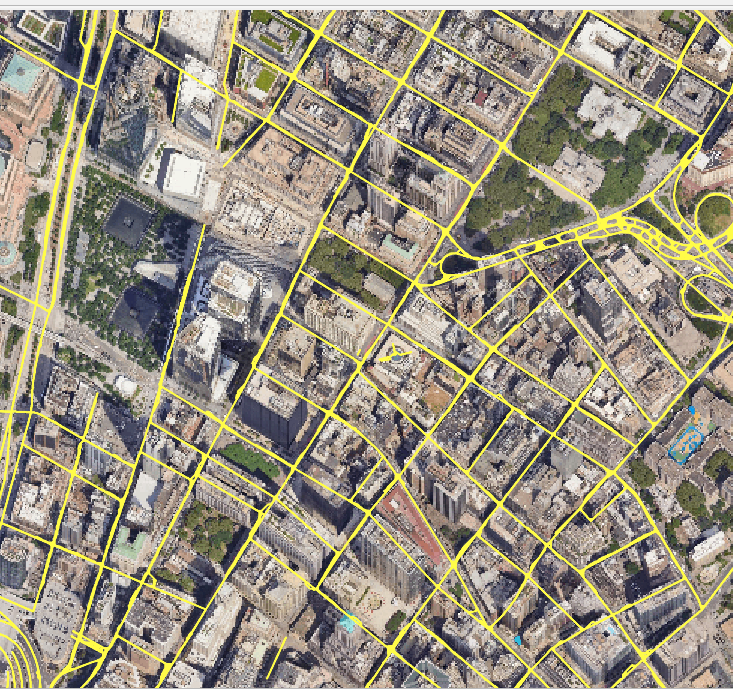}
        \caption{Final Output of IGraSS}
        \label{fig:final_output}
    \end{subfigure}
    \caption{Visualization of Road Network Output with IGraSS: The figure presents a comparison demonstrating the effectiveness of the IGraSS framework in repairing artificially introduced breaks in road network data. The top-left image displays a section of the original ground truth road network, providing a baseline for comparison. In the top-right image, we see the same network after introducing breaks of varying radius. The bottom image showcases the results of applying the IGraSS framework to this disrupted network. It demonstrates how IGraSS effectively reconstructing the road network. This visual comparison allows us to assess the performance of IGraSS in bridging gaps and restoring connectivity in road network data, highlighting its potential for improving the completeness and accuracy of road maps derived from imperfect or noisy input data.}
    \label{fig:road_viz}
\end{figure*}
\subsection{Parameter Sensitivity}
The IGraSS Framework incorporates several parameters: the number of intermediate epochs $E$, radius $\rho$, threshold $\tau$, and confidence threshold $\alpha$. Model performance varies depending on these parameters. Table \ref{tab:parameter_tune} presents comprehensive results of varying confidence thresholds and numbers of intermediate epochs. For these experiments, we used a radius $\rho = 100$ and reported average results from K-Fold Cross validation across validation sets. Comparing the results after 5 iterations, we observe that a threshold of 0.01 with 10 epochs outperforms others in terms of reducing unreachable canals and remaining terminals, but not in metric performance. Figure \ref{fig:error_results} visualizes the results for threshold 0.01 with 10 epochs, showing noisy precision leading to incorrect connections. In contrast, 20 epochs with a threshold of 0.1 after the first iteration produces cleaner results. Choosing too high a threshold might affect connections, as canal segments not present in the main ground truth might lead the model to assign lower confidence levels. In our case, thresholds between 0.2 and 0.1 worked well over iterations. The number of iterations is also crucial; after just 10 epochs with a lower threshold like 0.01, the model might predict considerable noise, whereas longer training might lead to learning the noisy ground truth. It may be beneficial to lower the threshold after the model has been trained for a sufficient number of epochs and has learned the correct features to connect more canals. In most of our analyses, we used confidence thresholds $\alpha$ ranging from 0.2 to 0.01 to train the model.

We conducted an additional set of parameter analyses by varying the radius $\rho$. Adjusting the radius can facilitate the connection of more gaps in the canal network. The optimal radius can be tuned according to the length of the gaps that need to be filled. As shown in Table \ref{tab:parameter}, compared to a radius of 20, more gaps were filled when using radii of 50, 100, and 150.

\begin{table*}[]
\caption{Parameter Analysis Varying intermediate epochs and confidence threshold $\alpha$}
\begin{tabular}{@{}c|c|c|ccccccccccc@{}}
\toprule
Model                        & $th$                  & \begin{tabular}[c]{@{}c@{}}E/\\ it\end{tabular} & P     & $\rp$ & R     & $\rr$ & F1    & rf    & D     & $\rd$ & \begin{tabular}[c]{@{}c@{}}R. \\ Nodes \%\end{tabular} & \begin{tabular}[c]{@{}c@{}}U. \\ Nodes \%\end{tabular} & $t$ \\ \midrule
\multirow{11}{*}{DeepLabv3+} & \multirow{3}{*}{0.3}  & 30                                              & 0.371 & 0.552 & 0.306 & 0.441              & 0.307 & 0.451 & 0.214 & 0.329              & 0.950                                                  & 0.0504                                                 & 202 \\
                             &                       & 20                                              & 0.447 & 0.654 & 0.305 & 0.432              & 0.348 & 0.503 & 0.241 & 0.366              & 0.939                                                  & 0.0615                                                 & 215 \\
                             &                       & 10                                              & 0.495 & 0.703 & 0.281 & 0.400              & 0.343 & 0.490 & 0.226 & 0.346              & 0.938                                                  & 0.0623                                                 & 227 \\ \cmidrule(l){2-14} 
                             & \multirow{2}{*}{0.2}  & 20                                              & 0.416 & 0.613 & 0.311 & 0.439              & 0.338 & 0.489 & 0.227 & 0.348              & 0.954                                                  & 0.0456                                                 & 176 \\
                             &                       & 10                                              & 0.507 & 0.716 & 0.270 & 0.384              & 0.337 & 0.480 & 0.226 & 0.345              & 0.943                                                  & 0.0574                                                 & 206 \\ \cmidrule(l){2-14} 
                             & \multirow{3}{*}{0.1}  & 30                                              & 0.416 & 0.621 & 0.254 & 0.371              & 0.291 & 0.434 & 0.213 & 0.330              & 0.965                                                  & 0.0353                                                 & 140 \\
                             &                       & 20                                              & 0.417 & 0.612 & 0.309 & 0.433              & 0.332 & 0.479 & 0.228 & 0.348              & 0.959                                                  & 0.0412                                                 & 155 \\
                             &                       & 10                                              & 0.502 & 0.708 & 0.276 & 0.391              & 0.339 & 0.482 & 0.227 & 0.346              & 0.956                                                  & 0.0437                                                 & 166 \\ \cmidrule(l){2-14} 
                             & \multirow{3}{*}{0.01} & 30                                              & 0.393 & 0.589 & 0.325 & 0.466              & 0.330 & 0.487 & 0.242 & 0.371              & 0.975                                                  & 0.0248                                                 & 91  \\
                             &                       & 20                                              & 0.396 & 0.587 & 0.292 & 0.422              & 0.300 & 0.442 & 0.215 & 0.333              & 0.975                                                  & 0.0250                                                 & 91  \\
                             &                       & 10                                              & 0.439 & 0.641 & 0.298 & 0.426              & 0.340 & 0.491 & 0.227 & 0.349              & 0.975                                                  & 0.0246                                                 & 89  \\ \midrule
\multirow{12}{*}{Resnet50}   & \multirow{3}{*}{0.3}  & 30                                              & 0.589 & 0.824 & 0.361 & 0.479              & 0.431 & 0.585 & 0.390 & 0.542              & 0.952                                                  & 0.0484                                                 & 191 \\
                             &                       & 20                                              & 0.503 & 0.733 & 0.428 & 0.573              & 0.450 & 0.629 & 0.395 & 0.569              & 0.953                                                  & 0.0471                                                 & 181 \\
                             &                       & 10                                              & 0.452 & 0.670 & 0.475 & 0.629              & 0.455 & 0.640 & 0.360 & 0.531              & 0.952                                                  & 0.0483                                                 & 188 \\ \cmidrule(l){2-14} 
                             & \multirow{3}{*}{0.2}  & 30                                              & 0.632 & 0.873 & 0.351 & 0.462              & 0.431 & 0.580 & 0.387 & 0.533              & 0.958                                                  & 0.0422                                                 & 169 \\
                             &                       & 20                                              & 0.572 & 0.811 & 0.372 & 0.494              & 0.433 & 0.594 & 0.377 & 0.533              & 0.953                                                  & 0.0471                                                 & 179 \\
                             &                       & 10                                              & 0.515 & 0.752 & 0.445 & 0.590              & 0.468 & 0.651 & 0.365 & 0.534              & 0.953                                                  & 0.0473                                                 & 181 \\ \cmidrule(l){2-14} 
                             & \multirow{3}{*}{0.1}  & 30                                              & 0.550 & 0.771 & 0.418 & 0.551              & 0.462 & 0.628 & 0.419 & 0.584              & 0.959                                                  & 0.0413                                                 & 163 \\
                             &                       & 20                                              & 0.555 & 0.797 & 0.413 & 0.545              & 0.461 & 0.634 & 0.400 & 0.568              & 0.958                                                  & 0.0425                                                 & 157 \\
                             &                       & 10                                              & 0.483 & 0.702 & 0.488 & 0.634              & 0.476 & 0.655 & 0.389 & 0.560              & 0.959                                                  & 0.0412                                                 & 153 \\ \cmidrule(l){2-14} 
                             & \multirow{3}{*}{0.01} & 30                                              & 0.602 & 0.846 & 0.344 & 0.459              & 0.417 & 0.571 & 0.372 & 0.522              & 0.968                                                  & 0.0322                                                 & 118 \\
                             &                       & 20                                              & 0.526 & 0.772 & 0.388 & 0.528              & 0.434 & 0.614 & 0.376 & 0.548              & 0.974                                                  & 0.0263                                                 & 100 \\
                             &                       & 10                                              & 0.473 & 0.698 & 0.477 & 0.622              & 0.465 & 0.647 & 0.377 & 0.548              & 0.976                                                  & 0.0243                                                 & 91  \\ \midrule
\multirow{12}{*}{ResUNet}    & \multirow{3}{*}{0.3}  & 30                                              & 0.580 & 0.820 & 0.322 & 0.427              & 0.388 & 0.531 & 0.313 & 0.449              & 0.946                                                  & 0.0631                                                 & 188 \\
                             &                       & 10                                              & 0.580 & 0.802 & 0.341 & 0.446              & 0.406 & 0.544 & 0.305 & 0.433              & 0.937                                                  & 0.0530                                                 & 219 \\
                             &                       & 20                                              & 0.586 & 0.819 & 0.351 & 0.460              & 0.417 & 0.562 & 0.320 & 0.457              & 0.949                                                  & 0.0590                                                 & 202 \\ \cmidrule(l){2-14} 
                             & \multirow{3}{*}{0.2}  & 30                                              & 0.611 & 0.850 & 0.278 & 0.372              & 0.354 & 0.484 & 0.280 & 0.402              & 0.951                                                  & 0.0504                                                 & 202 \\
                             &                       & 20                                              & 0.586 & 0.819 & 0.351 & 0.460              & 0.417 & 0.562 & 0.320 & 0.457              & 0.949                                                  & 0.0599                                                 & 202 \\
                             &                       & 10                                              & 0.587 & 0.806 & 0.320 & 0.420              & 0.385 & 0.515 & 0.288 & 0.410              & 0.940                                                  & 0.0508                                                 & 212 \\ \cmidrule(l){2-14} 
                             & \multirow{3}{*}{0.1} & 30                                              & 0.611 & 0.850 & 0.278 & 0.372              & 0.354 & 0.484 & 0.280 & 0.402              & 0.951                                                 & 0.0504                                                 & 203 \\
                             &                       & 20                                              & 0.614 & 0.848 & 0.321 & 0.422              & 0.398 & 0.536 & 0.306 & 0.436              & 0.950                                                  & 0.0491                                                 & 189 \\
                             &                       & 10                                              & 0.616 & 0.838 & 0.333 & 0.435              & 0.404 & 0.538 & 0.303 & 0.430              & 0.950                                                  & 0.0491                                                 & 189 \\ \cmidrule(l){2-14} 
                             & \multirow{3}{*}{0.01} & 30                                              & 0.612 & 0.842 & 0.335 & 0.457              & 0.417 & 0.571 & 0.372 & 0.522              & 0.968                                                 & 0.0324                                                 & 118 \\
                             &                       & 20                                              & 0.572 & 0.804 & 0.330 & 0.428              & 0.388 & 0.521 & 0.305 & 0.432              & 0.974                                                  & 0.0252                                                 & 91  \\
                             &                       & 10                                              & 0.585 & 0.801 & 0.309 & 0.412              & 0.376 & 0.509 & 0.277 & 0.398              & 0.974                                                  & 0.0252                                                 & 91  \\ \bottomrule
\end{tabular}
\label{tab:parameter}
\end{table*}

\subsection{Error Analysis}
In this section, we summarize potential errors that the IGraSS framework might introduce. As discussed in our parameter analysis, selecting appropriate parameters is crucial to avoid erroneous connections. IGraSS's focus on connecting points via the shortest path helps minimize errors when adding new data to the ground truth. Directly using the neural network output would have introduced significant noise to the ground truth, which our adaptive thresholding process helps mitigate. However, as illustrated in Figure \ref{fig:error_results_main}, unwanted connections may still occur. Additionally, the road network result visualization in Figure \ref{fig:road_viz} demonstrates that in areas with loops (visible in the right corner), the shortest path calculation may connect to side roads. While these are still roads, this issue arises due to width mismatches. Our train and test connectivity results reveal that some terminals remain unconnected. Figure \ref{fig:hardpredictions} visualizes data examples where certain pixels are exceptionally challenging for the model to predict, resulting in no connection being made by IGraSS.

\end{document}